\documentclass[preprint,12pt]{elsarticle}

\usepackage{xcolor,colortbl}
\usepackage{microtype}
\usepackage{graphicx}
\usepackage{subfigure}
\usepackage{caption}
\usepackage{booktabs} 
\usepackage{amsmath,amssymb,amsfonts}
\usepackage{tablefootnote}
\usepackage[hyphens,spaces,obeyspaces]{url}
\usepackage{hyperref}
\usepackage{multirow}

\DeclareMathAlphabet{\mathbbold}{U}{bbold}{m}{n}
    
\journal{Expert Systems with Applications}
\biboptions{authoryear}

\begin{document}

\begin{frontmatter}



\title{A Unified Benchmark for the Unknown Detection Capability of Deep Neural Networks}


\author[alpha]{Jihyo Kim}
\ead{jihyo.kim@ds.seoultech.ac.kr}
\author[alpha]{Jiin Koo}
\ead{jiinnine@ds.seoultech.ac.kr}
\author[alpha]{Sangheum Hwang\corref{cor1}}
\ead{shwang@seoultech.ac.kr}
\cortext[cor1]{Corresponding author}
\address[alpha]{Department of Data Science, Seoul National University of Science and Technology, \\
Seoul 01811, Republic of Korea}

\begin{abstract}
Deep neural networks have achieved outstanding performance over various tasks, but they have a critical issue: over-confident predictions even for completely unknown samples. Many studies have been proposed to successfully filter out these unknown samples, but they only considered narrow and specific tasks, referred to as misclassification detection, open-set recognition, or out-of-distribution detection. In this work, we argue that these tasks should be treated as fundamentally an identical problem because an ideal model should possess detection capability for all those tasks. Therefore, we introduce the unknown detection task, an integration of previous individual tasks, for a rigorous examination of the detection capability of deep neural networks on a wide spectrum of unknown samples. To this end, unified benchmark datasets on different scales were constructed and the unknown detection capabilities of existing popular methods were subject to comparison. We found that Deep Ensemble consistently outperforms the other approaches in detecting unknowns; however, all methods are only successful for a specific type of unknown. The reproducible code and benchmark datasets are available at \href{https://github.com/daintlab/unknown-detection-benchmarks}{https://github.com/daintlab/unknown-detection-benchmarks}.
\end{abstract}




\end{frontmatter}

\section{Introduction} \label{sec:introduction}
Deep neural networks have achieved significant performance improvements over various tasks such as image classification~\citep{krizhevsky2012imagenet, rawat2017deep}, object detection~\citep{ren2015fastrcnn, zhao2019object}, speech recognition~\citep{hinton2012deep, nassif2019speech}, and natural language processing~\citep{mikolov2010recurrent, devlin2019bert}. Despite these remarkable achievements, current deep neural networks have a critical deficiency that should be seriously considered before these models can be deployed in real-world applications. That is, they tend to make predictions with over-confidence even if the predictions are incorrect or the inputs are not relevant to a target task. This issue is a major concern in applications where models can cause fatal safety problems, such as medical diagnoses or autonomous driving~\citep{mehrtash2020medical, mohseni2020autonomous}. To establish reliable and secure systems with deep neural networks, our models should possess the important property of confidence in the predictions, described as: ``\emph{the model should know what it does not know.}'' Specifically, models should produce high confidence when encountering inputs likely to be correct predictions (i.e., known samples), whereas they should produce low confidence if predictions are likely to be incorrect or if inputs are semantically far from what was learned (i.e., unknown samples). Hence, it is necessary to have models that produce well-ranked confidence values according to how confident they are about the predictions. 

\begin{figure*}[!t]
\vskip 0.1in
    \centering
    \includegraphics[width=0.85\textwidth]{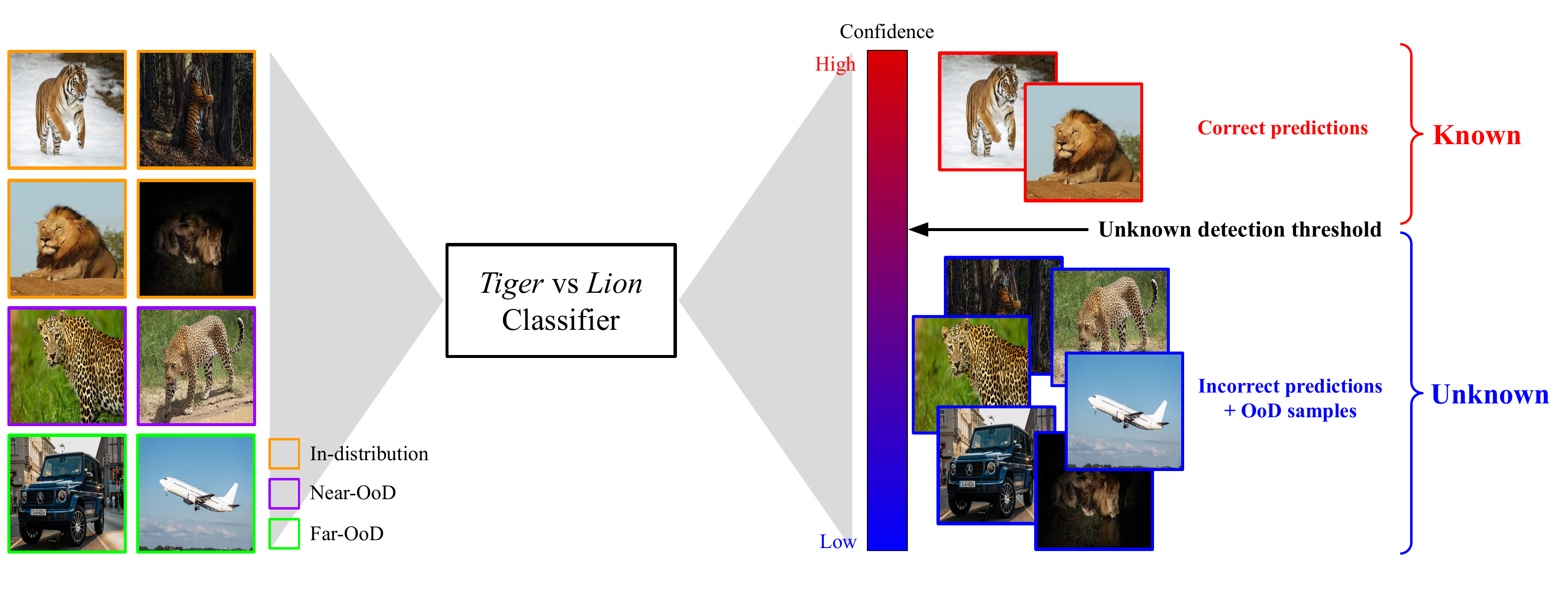}
    \caption{The overall concept of unknown detection. A model deployed in real-world applications may take not only inputs from the training distribution (in-distribution; orange boxes), but also those from distributions that are irrelevant to the target task (near-OoD; violet boxes, far-OoD; green boxes). For the model to be reliable, it should have the capability of distinguishing certainly known inputs (i.e., correct predictions) and unknown ones (i.e., incorrect predictions and OoD samples) based on the confidence score of each prediction.}
    \label{fig:concept}
\vskip -0.1in
\end{figure*}

Many works have sought to devise models that possess such property from two individual perspectives. First, previous researches named selective classification~\citep{el2010selective, geifman2017selective} or misclassification detection (MD) (i.e., failure prediction)~\citep{jiang2018trust, corbiere2019addressing} set the objective to provide a well-aligned ordinal ranking of confidence values to clarify the separation between correct and incorrect predictions on samples from a single data distribution (i.e., in-distribution).
Another line of research considered the problem of open-set recognition (OsR) or out-of-distribution detection (OoDD), which aim to estimate well-ranked confidence values between samples from the in-distribution and those from other distributions, referred to as out-of-distribution~\citep{bendale2016towards, liang2017enhancing, lee2018simple}. In other words, the former focuses on the ranking of confidence values within the in-distribution, whereas the latter takes into account that between the in-distribution and out-of-distributions. We argue that these two separate views should be integrated to evaluate rigorously the true detection capability of deep neural networks on unknown inputs, as unknowns should include both incorrectly predicted in-distribution samples and unpredictable out-of-distribution samples. It should be noted that our definition of unknowns refers to samples, not classes. This means that even if a test input belongs to one of the in-distribution classes, it would still be considered an unknown sample to a classifier if it leads to a false prediction.

Those different perspectives have led to individual benchmark settings, although previous works have attempted to solve a fundamentally similar issue~\citep{moon2020confidence}. MD considers the setting assuming that all inputs at test time come from the in-distribution. In other words, previous works for MD do not take into account inputs lying outside of the training data distribution. Therefore, a single dataset such as CIFAR-10/100~\citep{krizhevsky2009learning} is sufficient to evaluate the MD performance of a model~\citep{geifman2018bias, corbiere2020confidence}. OsR and OoDD can be categorized as the same problem; however, the benchmark settings are slightly different in the literature. Typically, OoDD considers OoD as a completely different distribution from the training distribution~\citep{lee2017training}, and therefore, focuses on detecting inputs that are semantically far from the in-distribution dataset. For instance, SVHN~\citep{Netzer2011svhn}, LSUN~\citep{yu2015lsun} or synthetic data such as Gaussian noise are frequently used as OoD datasets when CIFAR-10 is the in-distribution dataset~\citep{liang2017enhancing, lee2018simple}. Some previous works focusing on OsR were undertaken in settings similar to those of OoDD~\citep{oza2019cnnosr, perera2020genosr}, but a few works set OoD as a semantically close distribution~\citep{jang2020onerest, oza2019c2ae}, e.g., images containing the same object but from other datasets~\citep{perera2020genosr}. These individually different benchmark settings make it difficult to compare existing methods proposed for basically similar problems, i.e., MD, OsR, and OoDD.

We thus introduce a more fundamental task, termed \emph{unknown detection}, which integrates the problems of MD, OsR, and OoDD, independently studied in the literature. The goal of unknown detection is to make a neural network model that produces well-aligned confidence scores between inputs likely to be predicted correctly and those regarded as unknowns, including inputs likely predicted incorrectly and not relevant to the task of the model. If a model produces such well-aligned confidence scores, it can be successfully deployed for safety-critical applications, as the system can request user intervention when the confidence score of the input is below a predefined threshold.

Figure~\ref{fig:concept} depicts the overall concept of our unknown detection problem. Suppose that a practitioner wants to deploy the tiger vs. lion classifier in a real-world application. In this scenario, the classifier may take inputs containing various objects that can be classified into three categories: \romannumeral 1 ) target objects, e.g., a \texttt{lion} or a \texttt{tiger}; \romannumeral 2 ) objects semantically similar to the target objects (i.e., near-OoD), such as a \texttt{leopard}; and \romannumeral 3 ) objects completely different from the target objects (i.e., far-OoD), e.g., an \texttt{automobile} or an \texttt{airplane}. If the model can produce predictions with well-aligned confidence values as depicted on the right side of Figure~\ref{fig:concept}, it implies that the model has the capability to distinguish correctly predicted inputs (red boxes) from others including incorrect predictions, near- and far-OoD (blue boxes) when the confidence threshold is properly set. Therefore, it can be utilized with a high degree of reliability for safety-critical systems.

To examine the unknown detection capability of deep neural networks, we propose new benchmark settings based on popular datasets, such as CIFAR-100 and ImageNet~\citep{deng2009imagenet}, which cover the complete spectrum of the unknown detection problem. Specifically, our benchmark setting consists of three categories of datasets: in-distribution, near-OoD, and far-OoD. Near- and far-OoD are designed to be distinguished according to how the distribution is semantically overlapped with the in-distribution; i.e., near-OoD shares some degree of semantic concepts with the in-distribution dataset, whereas far-OoD contains completely different semantics. In this work, the concept of the superclass is utilized to distinguish near- and far-OoD classes given in-distribution classes. For example, the given superclass structure of CIFAR-100 was used as a criterion to choose near-OoD classes for the proposed CIFAR-100-based benchmark. In the ImageNet-based benchmark, the WordNet~\citep{miller1995wordnet} tree structure was used, as all classes in ImageNet inherit the hierarchical structure of WordNet. The distance between class keywords in WordNet was computed to determine whether classes are semantically similar (i.e., near-OoD) or very different (i.e., far-OoD).

With the proposed benchmarks, we evaluate the unknown detection capability of several popular methods proposed for MD, OsR, and OoDD. Our experimental results reveal that Deep Ensemble~\citep{lakshminarayanan2016simple} is most competitive in terms of unknown detection, although there is still room for improvement in terms of OoDD. Surprisingly, we found that existing methods for a particular task do not work well on other tasks. For example, the effective methods for MD do not show good performance on OoDD; on the other hand, popular methods for OoDD can even harm the MD task.

Our contributions are summarized as follows:
\begin{itemize}
    \item We suggest the new task \emph{unknown detection} which integrates MD, OsR, and OoDD, to evaluate the true detection capability of unknown inputs. Its goal is to detect correctly predicted inputs (i.e., known inputs) against incorrectly predicted inputs as well as task-irrelevant inputs (i.e., unknown inputs).
    \item We propose unified benchmarks for evaluating the unknown detection capability of a neural network model; these consist of in-distribution, near-OoD, and far-OoD datasets. 
    \item We evaluate the unknown detection capabilities of popular methods for the tasks of MD, OsR, and OoDD with the proposed benchmarks. Although Deep Ensemble beats other comparative approaches in terms of unknown detection, we found that it still has room for improvement when it comes to detecting OoDD.
\end{itemize}

\section{Related Work} \label{sec:seection2}
A variety of methods for each task, MD, OsR, and OoDD, have been actively proposed to detect test inputs that are incorrectly predicted (i.e., MD) or not relevant to a target task (i.e., OsR or OoDD) in the context of image classification. Previous methods were developed under a specific problem setting, an incomplete approach from the perspective of unknown detection, and were evaluated on individual task-specific benchmarks accordingly as summarized in Table~\ref{table:benchmark-setting}.

\textbf{Misclassification detection (MD).} 
MD aims to align the ordinal ranking of confidence values well to clarify the separation between correct and incorrect predictions of unseen samples from the training distribution. That is, MD assumes that the test inputs are the \textit{i.i.d.} samples of the training distribution. Essentially, predicted class probabilities via a softmax function can be used for detecting samples likely to be classified incorrectly. \cite{hendrycks2016baseline} suggested exploiting the softmax outputs of conventional deep classifiers to separate correctly and incorrectly classified samples, demonstrating that this simple approach can achieve relatively good performance on MD. A selective classifier abstains the predictions of samples whose confidence values are below a threshold such that the error rate set by a user can be achieved while having an optimal coverage rate~\citep{geifman2017selective}.
\cite{geifman2018bias} observed that neural networks produce overconfident predictions especially for easy samples (i.e., correctly classified at the early stage of training) as training proceeds. To mitigate this issue, they proposed using earlier snapshots of the trained model and showed that it improves the quality of the confidence values in terms of the area under the risk-coverage curve (AURC). \cite{corbiere2019addressing} and \cite{moon2020confidence} proposed the direct learning of the true class probability and the ordinal ranking of confidence values, respectively.

\begin{table}[!t]
    \centering
    \caption{Benchmark settings and target problems in the literature on MD, OsR, or OoDD.}
    \label{table:benchmark-setting}
    \resizebox{0.5\textwidth}{!}{%
    \begin{tabular}{c|ccc|ccc}
    \toprule
    & \multicolumn{3}{c|}{\textbf{Benchmark settings}} & \textbf{MD} & \begin{tabular}[c]{@{}c@{}}\textbf{Near} \\ \textbf{OoD}\end{tabular} & \begin{tabular}[c]{@{}c@{}}\textbf{Far} \\ \textbf{OoD}\end{tabular} \\ \midrule
    
    \textbf{\cite{corbiere2019addressing}} & \multicolumn{3}{c|}{MNIST, SVHN, CIFAR-10/-100} & \checkmark & - &  - \\ \midrule
    \textbf{\cite{geifman2018bias}} & \multicolumn{3}{c|}{SVHN, CIFAR-10/-100, ImageNet} & \checkmark & - &  - \\ \midrule
    \multirow{2}{*}{\textbf{\cite{sensoy2018edl}}} & \multicolumn{3}{c|}{MNIST, CIFAR-10} & \checkmark & - & - \\ 
    & CIFAR-10 & \large{vs} & CIFAR-10 & - & \checkmark & - \\ \midrule
    \multirow{4}{*}{\textbf{\cite{moon2020confidence}}} & \multicolumn{3}{c|}{SVHN, CIFAR-10, CIFAR-100} & \checkmark & - &  - \\ 
    & \begin{tabular}[c]{@{}c@{}} SVHN\\CIFAR-10 \end{tabular} & \large{vs} & \begin{tabular}[c]{@{}c@{}} Tiny ImageNet\\LSUN\\iSUN \end{tabular} & - & - & \checkmark \\ \midrule
    \textbf{\cite{lee2018simple}} & CIFAR-10/-100 & \large{vs} & \begin{tabular}[c]{@{}c@{}} Tiny ImageNet\\LSUN\\SVHN \end{tabular} & - & - & \checkmark \\ \midrule
    \textbf{\cite{liang2017enhancing}} & CIFAR-10/-100 & \large{vs} & \begin{tabular}[c]{@{}c@{}} Tiny ImageNet\\LSUN\\iSUN\\Gaussian \end{tabular} & - & - &  \checkmark \\ \midrule
    \textbf{\cite{hendrycks2018deep}} & \begin{tabular}[c]{@{}c@{}}CIFAR-10/-100  \\ Tiny ImageNet \end{tabular}& \large{vs} & \begin{tabular}[c]{@{}c@{}} CIFAR-100/-10\\LSUN\\SVHN\\ImageNet\\Places365\\DTD\\Gaussian \end{tabular} & - & - & \checkmark \\ \midrule
    \textbf{\cite{oza2019cnnosr, oza2019c2ae}} & \begin{tabular}[c]{@{}c@{}} SVHN \\ CIFAR-10 \\ Tiny ImageNet \end{tabular}& \large{vs} & \begin{tabular}[c]{@{}c@{}} SVHN \\ CIFAR-10/-100 \\ Tiny ImageNet \end{tabular} & - & \checkmark & \checkmark \\ \midrule
    \textbf{\cite{perera2020genosr}} & CIFAR-10 & \large{vs} & CIFAR-100 & - & \checkmark & - \\ \midrule
    \textbf{\cite{sun2020conditional}} & \begin{tabular}[c]{@{}c@{}} CIFAR-10 \\ Tiny ImageNet \end{tabular} & \large{vs} & \begin{tabular}[c]{@{}c@{}} CIFAR-100 \\ Tiny ImageNet \end{tabular} & - & \checkmark & - \\ \midrule
    \textbf{\cite{jang2020onerest}} & CIFAR-100 & \large{vs} & CIFAR-100 & - & \checkmark & - \\
    \bottomrule
    \end{tabular}%
    }
    \vskip -0.1in
\end{table}

\textbf{Out-of-distribution detection (OoDD) (a.k.a. Open-set recognition (OsR)).}\footnote{Hereafter, we use the term OoDD only as the problem of OsR is fundamentally identical to that of OoDD.} 
The goal of OoDD is to ensure that a model provides well-aligned confidence values that clearly distinguish between in-distribution inputs and OoD inputs. \cite{hendrycks2016baseline} demonstrated that deep classifiers with softmax outputs are also effective for OoDD along with MD, as noted earlier. It was also found that post-processing techniques such as temperature scaling or input perturbation can effectively enhance the OoDD performance of standard classifiers~\citep{liang2017enhancing, lee2018simple}. To produce low confidence values for OoD inputs, several studies have suggested modifying the model architecture, such as by adding a confidence estimation branch~\citep{devries2018learning}. Intuitively, exploiting OoD samples explicitly during training can boost the OoDD performance of a model. In the literature, such OoD samples for training (also called external samples) consist of natural images~\citep{hendrycks2018deep} or generated synthetic images~\citep{lee2017training, vernekar2019genood}. With regard to synthetic images, they can be generated by a generative adversarial network (GAN)~\citep{lee2017training, neal2018openset, ge2017gopenmax} or a conditional variational auto-encoder~\citep{vernekar2019genood}. Extreme value theory can also be employed to update softmax scores for detecting OoD inputs. \cite{bendale2016towards} introduced the OpenMax layer to estimate the softmax values of inputs from an unknown distribution using extreme value theory. Several studies~\citep{oza2019c2ae, oza2019cnnosr} have utilized auto-encoder architectures with extreme value theory.

Other studies attempted to solve OoDD under an unsupervised learning setting. A likelihood provided by a generative model (e.g., PixelCNN++~\citep{salimans2017pixelcnn}, Glow~\citep{kingma2018glow}) or a hybrid model~\citep{nalisnick2019hybrid} can be used as a criterion to distinguish in-distribution inputs from OoD inputs~\citep{ren2019likelihood}. However, it is known that such likelihood-based approaches tend to assign a high likelihood to what are clearly OoD inputs~\citep{choi2019waic,nalisnick2018know}. Contrastive learning~\citep{chen2020simple}, a recently emerging method of self-supervised learning, also improves the OoDD performance~\citep{winkens2020contrastive}.

With regard to benchmark settings, it is most common to set datasets that are completely different from the in-distribution dataset as the OoD datasets~\citep{liang2017enhancing, lee2018simple, ge2017gopenmax}. However, not all studies share such settings, even when they focus on the same objective of OoDD. A few studies~\citep{oza2019cnnosr, perera2020genosr, neal2018openset} construct the in-distribution dataset and OoD dataset by selecting only specific classes from candidate datasets to consider a notion of semantic similarity, for example, non-animal classes from CIFAR-10 vs. animal classes from CIFAR-100~\citep{oza2019cnnosr}, or vehicle classes from CIFAR-10 vs. vehicle classes from CIFAR-100~\citep{perera2020genosr}. On the other hand, other studies~\citep{jang2020onerest, sun2020conditional} create an in-distribution dataset and an OoD dataset by intra-dataset division. \cite{roady2020large} also pointed out the problem of individual benchmark settings in the OoDD literature and compared several methods proposed for OoDD under the same benchmark setting. \cite{tajwar2021no} empirically demonstrated that none of the OoDD methods consistently outperforms others on a standardized OoDD setting. However, the comparison is limited to in-distribution vs. far-OoD, which merely covers a narrow scope of our unknown detection problem.

Overall, previous works on OoDD have assumed that unknown inputs at test time come from entirely different distributions relative to the in-distribution. However, this assumption does not consider the distinguishability between correct predictions and incorrect predictions of the in-distribution samples. Because good detection capability of OoD inputs does not guarantee good detection capability of incorrect predictions as confirmed in our experiments, one cannot ensure that a model produces well-ranked confidence values in terms of unknown detection.

\section{Unknown Detection Benchmark Datasets} \label{sec:section3}
In this section, we present the detailed procedure of how to construct benchmark datasets for the unknown detection task. Specifically, we design two benchmark datasets, the CIFAR- and ImageNet-based datasets, to consider different data scales in terms of the number of samples and the image resolution. 

As summarized in Table~\ref{table:benchmark-setting}, existing studies have focused on only a subset of the unknown detection problem, e.g., MD or OoDD tasks, and thus have been individually evaluated on their own benchmark settings. However, all detection targets, including incorrect predictions in MD and near-/far-OoD inputs in OoDD, should be taken into consideration when we define the unknown samples for a model. Thus, our unknown detection benchmark datasets basically consist of the following categories: in-distribution, near-OoD, and far-OoD datasets. 

The in-distribution dataset is used to train a model. Given that the model barely achieves perfect accuracy, test samples from this dataset can be classified into correct (i.e., known) and incorrect (i.e., unknown) predictions, as in the typical setting of the MD problem. The OoD dataset in our benchmark consists of near- and far-OoD datasets that are used to evaluate the detection capability on inputs not relevant to a given task. Here, we employ the semantic distance between classes as the notion of the distance. More concretely, a specific class is regarded as the near-OoD class if there is at least one class in the in-distribution dataset that shares a superclass. For example, if two classes A and B belong to a class C (i.e., C is a superclass of A and B), then B is considered as a near-OoD class of A. Otherwise, it is considered as the far-OoD class if it comes from totally different sources (e.g., digits vs. CIFAR classes).

\begin{table}[!t]
    \centering
    \caption{Summary of the proposed benchmark datasets. Note that the external training and validation datasets are used for some methods that require OoD samples for model training (e.g., OE) and for hyperparameter search (e.g., ODIN), respectively.}
    \label{table:benchmark-summary}
    \resizebox{0.5\textwidth}{!}{%
    \begin{tabular}{l|c|ccc}
    \toprule
    
    & \textbf{Dataset} & \textbf{Train} &\textbf{Validation} & \textbf{Test} \\ \midrule
    
    \multirow{13}{*}{\rotatebox[origin=c]{90}{\textbf{CIFAR}}} & \cellcolor{gray!20}In-distribution Dataset & & & \\
    & CIFAR-40 & 18,000 & 2,000 & 4,000 \\ \cmidrule{2-5}
    & \cellcolor{gray!20}External Dataset & & & \\ 
    & Tiny ImageNet158-FIX & 79,000 & 2,000 & \large{-} \\  \cmidrule{2-5}
    & \cellcolor{gray!20}Near-OoD Dataset & & & \\ 
    & CIFAR-60 & \large{-} & \large{-} & 6,000 \\ \cmidrule{2-5}
    & \cellcolor{gray!20}Far-OoD Dataset & & & \\
    & LSUN-FIX & \large{-} & \large{-} & 4,000 \\
    & SVHN& \large{-} & \large{-} & 4,000 \\
    & DTD & \large{-} & \large{-} & 5,640 \\
    & Gaussian & \large{-} & \large{-} & 10,000 \\ \midrule
    
    \multirow{11}{*}{\rotatebox[origin=c]{90}{\textbf{ImageNet}}} & \cellcolor{gray!20}In-distribution Dataset & & & \\
    & ImageNet-200 & 250,000 & 10,000 & 10,000 \\ \cmidrule{2-5}
    & \cellcolor{gray!20}External Dataset & & & \\ 
    & External ImageNet-394 & 492,500 & 19,700 & \large{-} \\  \cmidrule{2-5}
    & \cellcolor{gray!20}Near-OoD Dataset & & & \\ 
    & Near ImageNet-200 & \large{-} & \large{-} & 10,000 \\ \cmidrule{2-5}
    & \cellcolor{gray!20}Far-OoD Dataset & & & \\
    & Food-32 & \large{-} & \large{-} & 10,000 \\
    & Caltech-45& \large{-} & \large{-} & 4,792 \\
    & Places-82 & \large{-} & \large{-} & 10,000 \\ \bottomrule
    \end{tabular}%
    }
    \vskip -0.1in
\end{table}

One more dataset is constructed to compare the existing methods in a fair setting: the external dataset. This dataset is used for hyperparameter tuning or the training of certain methods if they require an auxiliary dataset such as OE~\citep{hendrycks2018deep}. Note that some existing works select the optimal hyperparameters using randomly sampled data from the target OoD. For these methods, the external dataset is used for hyperparameter tuning, as the aforementioned scenario is practically infeasible. Importantly, samples from the in-distribution dataset should not be semantically overlapped with those from far-OoD or an external dataset so as to evaluate the unknown detection capability rigorously. For example, it is difficult to evaluate the performance accurately or to find optimal hyperparameters if OoD classes or external dataset classes are semantically similar to those in the in-distribution dataset. To this end, we employ the WordNet tree to measure the semantic distance between classes approximately. \cite{pedersen2004wordnet} employed the least common subsumer and path length to measure the similarity between two concepts in WordNet. Our approaches to measuring the semantic distance between classes are inspired by their similarity measures. According to this approach, we eliminated several classes in the far-OoD dataset or the external dataset to minimize the semantic overlap with the in-distribution classes.

The following sections describe the construction processes of the proposed benchmark datasets based on CIFAR-100 and ImageNet for the unknown detection task. Table~\ref{table:benchmark-summary} summarizes the composition of the constructed unified benchmark datasets.

\begin{table}[!t]
    \centering
    \caption{All superclasses and classes belonging to CIFAR-100. Classes shown in \textcolor{green}{green} and \textcolor{yellow}{yellow} compose CIFAR-40 and CIFAR-60, respectively.}
    \label{table:cifar-split}
    \vskip 0.1in 
    \resizebox{0.5\textwidth}{!}{%
    \begin{tabular}{l|ccccc}
    \toprule
    \textbf{\large{Superclass}} & & & \textbf{\large{Classes}} & & \\
    
    \midrule
    
    \cellcolor{gray!20} & \cellcolor{yellow!50} & \cellcolor{yellow!50} & \cellcolor{green!30} & \cellcolor{yellow!50} & \cellcolor{green!30} \\
    \cellcolor{gray!20} & \cellcolor{yellow!50}\large{beaver} & \cellcolor{yellow!50}\large{dolphin} & \cellcolor{green!30}\large{otter} & \cellcolor{yellow!50}\large{seal} & \cellcolor{green!30}\large{whale} \\
    \multirow{-3}{*}{\cellcolor{gray!20}\large{\begin{tabular}[l]{@{}l@{}} aquatic \\ mammals \end{tabular}}} & \cellcolor{yellow!50} & \cellcolor{yellow!50} & \cellcolor{green!30} & \cellcolor{yellow!50} & \cellcolor{green!30} \\
    
    \multirow{3}{*}{\large{fish}} & \cellcolor{green!30} & \cellcolor{yellow!50} & \cellcolor{green!30} & \cellcolor{yellow!50} & \cellcolor{yellow!50} \\
    & \cellcolor{green!30}\large{aquarium fish} & \cellcolor{yellow!50}\large{flatfish} & \cellcolor{green!30}\large{ray} & \cellcolor{yellow!50}\large{shark} & \cellcolor{yellow!50}\large{trout} \\
    & \cellcolor{green!30} & \cellcolor{yellow!50} & \cellcolor{green!30} & \cellcolor{yellow!50} & \cellcolor{yellow!50} \\
    
    \cellcolor{gray!20} & \cellcolor{green!30} & \cellcolor{green!30} & \cellcolor{yellow!50} & \cellcolor{yellow!50} & \cellcolor{yellow!50} \\
    \cellcolor{gray!20} & \cellcolor{green!30}\large{orchids} & \cellcolor{green!30}\large{poppies} & \cellcolor{yellow!50}\large{roses}  & \cellcolor{yellow!50}\large{sunflowers} & \cellcolor{yellow!50}\large{tulips} \\
    \multirow{-3}{*}{\cellcolor{gray!20}\large{flowers}} & \cellcolor{green!30} & \cellcolor{green!30} & \cellcolor{yellow!50} & \cellcolor{yellow!50} & \cellcolor{yellow!50} \\
    
    \multirow{3}{*}{\large{\begin{tabular}[l]{@{}l@{}} food \\ containers \end{tabular}}} & \cellcolor{yellow!50} & \cellcolor{yellow!50} & \cellcolor{green!30} & \cellcolor{yellow!50} & \cellcolor{green!30} \\
    & \cellcolor{yellow!50}\large{bottles} & \cellcolor{yellow!50}\large{bowls} & \cellcolor{green!30}\large{cans} & \cellcolor{yellow!50}\large{cups} & \cellcolor{green!30}\large{plates} \\
    & \cellcolor{yellow!50} & \cellcolor{yellow!50} & \cellcolor{green!30} & \cellcolor{yellow!50} & \cellcolor{green!30} \\
    
    \cellcolor{gray!20} & \cellcolor{yellow!50} & \cellcolor{yellow!50} & \cellcolor{green!30} & \cellcolor{yellow!50} & \cellcolor{green!30} \\
    \cellcolor{gray!20} & \cellcolor{yellow!50}\large{apples} & \cellcolor{yellow!50}\large{mushrooms} & \cellcolor{green!30}\large{oranges} & \cellcolor{yellow!50}\large{pears} & \cellcolor{green!30}\large{sweet peppers} \\
    \multirow{-3}{*}{\cellcolor{gray!20}\large{\begin{tabular}[l]{@{}l@{}} fruit and \\ vegetables \end{tabular}}} & \cellcolor{yellow!50} & \cellcolor{yellow!50} & \cellcolor{green!30} & \cellcolor{yellow!50} & \cellcolor{green!30} \\
    
    \multirow{3}{*}{\large{\begin{tabular}[l]{@{}l@{}} household \\ electrical devices \end{tabular}}} & \cellcolor{green!30} & \cellcolor{yellow!50} & \cellcolor{yellow!50} & \cellcolor{yellow!50} & \cellcolor{green!30} \\
    & \cellcolor{green!30}\large{clock} & \cellcolor{yellow!50}\large{computer keyboard} & \cellcolor{yellow!50}\large{lamp} & \cellcolor{yellow!50}\large{telephone} & \cellcolor{green!30}\large{television} \\
    & \cellcolor{green!30} & \cellcolor{yellow!50} & \cellcolor{yellow!50} & \cellcolor{yellow!50} & \cellcolor{green!30} \\
    
    \cellcolor{gray!20} & \cellcolor{green!30} & \cellcolor{green!30} & \cellcolor{yellow!50} & \cellcolor{yellow!50} & \cellcolor{yellow!50} \\
    \cellcolor{gray!20} & \cellcolor{green!30}\large{bed} & \cellcolor{green!30}\large{chair} & \cellcolor{yellow!50}\large{couch} & \cellcolor{yellow!50}\large{table} & \cellcolor{yellow!50}\large{wardrobe} \\
    \multirow{-3}{*}{\cellcolor{gray!20}\large{\begin{tabular}[l]{@{}l@{}} household \\ furniture \end{tabular}}} & \cellcolor{green!30} & \cellcolor{green!30} & \cellcolor{yellow!50} & \cellcolor{yellow!50} & \cellcolor{yellow!50} \\
    
    \multirow{3}{*}{\large{insects}} & \cellcolor{yellow!50} & \cellcolor{green!30} & \cellcolor{yellow!50} & \cellcolor{green!30} & \cellcolor{yellow!50} \\
    & \cellcolor{yellow!50}\large{bee} & \cellcolor{green!30}\large{beetle} & \cellcolor{yellow!50}\large{butterfly} & \cellcolor{green!30}\large{caterpillar} & \cellcolor{yellow!50}\large{cockroach} \\
    & \cellcolor{yellow!50} & \cellcolor{green!30} & \cellcolor{yellow!50} & \cellcolor{green!30} & \cellcolor{yellow!50} \\
    
    \cellcolor{gray!20} & \cellcolor{yellow!50} & \cellcolor{yellow!50} & \cellcolor{yellow!50} & \cellcolor{green!30} & \cellcolor{green!30} \\
    \cellcolor{gray!20} & \cellcolor{yellow!50}\large{bear} & \cellcolor{yellow!50}\large{leopard} & \cellcolor{yellow!50}\large{lion} & \cellcolor{green!30}\large{tiger} & \cellcolor{green!30}\large{wolf} \\
    \multirow{-3}{*}{\cellcolor{gray!20}\large{\begin{tabular}[l]{@{}l@{}} large \\ carnivores \end{tabular}}} & \cellcolor{yellow!50} & \cellcolor{yellow!50} & \cellcolor{yellow!50} & \cellcolor{green!30} & \cellcolor{green!30} \\
    
    \multirow{3}{*}{\large{\begin{tabular}[l]{@{}l@{}} large man-made \\ outdoor things \end{tabular}}} & \cellcolor{yellow!50} & \cellcolor{green!30} & \cellcolor{yellow!50} & \cellcolor{green!30} & \cellcolor{yellow!50} \\
    & \cellcolor{yellow!50}\large{bridge} & \cellcolor{green!30}\large{castle} & \cellcolor{yellow!50}\large{house} & \cellcolor{green!30}\large{road} & \cellcolor{yellow!50}\large{skyscraper} \\
    & \cellcolor{yellow!50} & \cellcolor{green!30} & \cellcolor{yellow!50} & \cellcolor{green!30} & \cellcolor{yellow!50} \\
    
    \cellcolor{gray!20} & \cellcolor{yellow!50} & \cellcolor{green!30} & \cellcolor{yellow!50} & \cellcolor{yellow!50} & \cellcolor{green!30} \\
    \cellcolor{gray!20} & \cellcolor{yellow!50}\large{cloud} & \cellcolor{green!30}\large{forest} & \cellcolor{yellow!50}\large{mountain} & \cellcolor{yellow!50}\large{plain} & \cellcolor{green!30}\large{sea} \\
    \multirow{-3}{*}{\cellcolor{gray!20}\large{\begin{tabular}[l]{@{}l@{}} large natural \\ outdoor scenes \end{tabular}}} & \cellcolor{yellow!50} & \cellcolor{green!30} & \cellcolor{yellow!50} & \cellcolor{yellow!50} & \cellcolor{green!30} \\
    
    \multirow{3}{*}{\large{\begin{tabular}[l]{@{}l@{}} large omnivores \\ and herbivores \end{tabular}}} & \cellcolor{green!30} & \cellcolor{yellow!50} & \cellcolor{green!30} & \cellcolor{yellow!50} & \cellcolor{yellow!50} \\
    & \cellcolor{green!30}\large{camel} & \cellcolor{yellow!50}\large{cattle} & \cellcolor{green!30}\large{chimpanzee} & \cellcolor{yellow!50}\large{elephant} & \cellcolor{yellow!50}\large{kangaroo} \\
    & \cellcolor{green!30} & \cellcolor{yellow!50} & \cellcolor{green!30} & \cellcolor{yellow!50} & \cellcolor{yellow!50} \\
    
    \cellcolor{gray!20} & \cellcolor{yellow!50} & \cellcolor{yellow!50} & \cellcolor{green!30} & \cellcolor{green!30} & \cellcolor{yellow!50} \\
    \cellcolor{gray!20} & \cellcolor{yellow!50}\large{fox} & \cellcolor{yellow!50}\large{porcupine} & \cellcolor{green!30}\large{possum} & \cellcolor{green!30}\large{raccoon} & \cellcolor{yellow!50}\large{skunk} \\
    \multirow{-3}{*}{\cellcolor{gray!20}\large{\begin{tabular}[l]{@{}l@{}} medium-sized \\ mammals \end{tabular}}} & \cellcolor{yellow!50} & \cellcolor{yellow!50} & \cellcolor{green!30} & \cellcolor{green!30} & \cellcolor{yellow!50} \\
    
    \multirow{3}{*}{\large{\begin{tabular}[l]{@{}l@{}} non-insect \\ invertebrates \end{tabular}}} & \cellcolor{yellow!50} & \cellcolor{green!30} & \cellcolor{green!30} & \cellcolor{yellow!50} & \cellcolor{yellow!50} \\
    & \cellcolor{yellow!50}\large{crab} & \cellcolor{green!30}\large{lobster} & \cellcolor{green!30}\large{snail} & \cellcolor{yellow!50}\large{spider} & \cellcolor{yellow!50}\large{worm} \\
    & \cellcolor{yellow!50} & \cellcolor{green!30} & \cellcolor{green!30} & \cellcolor{yellow!50} & \cellcolor{yellow!50} \\
    
    \cellcolor{gray!20} & \cellcolor{yellow!50} & \cellcolor{yellow!50} & \cellcolor{green!30} & \cellcolor{green!30} & \cellcolor{yellow!50} \\
    \cellcolor{gray!20} & \cellcolor{yellow!50}\large{baby} & \cellcolor{yellow!50}\large{boy} & \cellcolor{green!30}\large{girl} & \cellcolor{green!30}\large{man} & \cellcolor{yellow!50}\large{woman} \\
    \multirow{-3}{*}{\cellcolor{gray!20}\large{people}} & \cellcolor{yellow!50} & \cellcolor{yellow!50} & \cellcolor{green!30} & \cellcolor{green!30} & \cellcolor{yellow!50} \\
    
    \multirow{3}{*}{\large{reptiles}} & \cellcolor{green!30} & \cellcolor{yellow!50} & \cellcolor{yellow!50} & \cellcolor{yellow!50} & \cellcolor{green!30} \\
    & \cellcolor{green!30}\large{crocodile} & \cellcolor{yellow!50}\large{dinosaur} & \cellcolor{yellow!50}\large{lizard} & \cellcolor{yellow!50}\large{snake} & \cellcolor{green!30}\large{turtle} \\
    & \cellcolor{green!30} & \cellcolor{yellow!50} & \cellcolor{yellow!50} & \cellcolor{yellow!50} & \cellcolor{green!30} \\
    
    \cellcolor{gray!20} & \cellcolor{yellow!50} & \cellcolor{green!30} & \cellcolor{yellow!50} & \cellcolor{yellow!50} & \cellcolor{green!30} \\
    \cellcolor{gray!20} & \cellcolor{yellow!50}\large{hamster} & \cellcolor{green!30}\large{mouse} & \cellcolor{yellow!50}\large{rabbit} & \cellcolor{yellow!50}\large{shrew} & \cellcolor{green!30}\large{squirrel} \\
    \multirow{-3}{*}{\cellcolor{gray!20}\large{\begin{tabular}[l]{@{}l@{}} small \\ mammals \end{tabular}}} & \cellcolor{yellow!50} & \cellcolor{green!30} & \cellcolor{yellow!50} & \cellcolor{yellow!50} & \cellcolor{green!30} \\
    
    \multirow{3}{*}{\large{trees}} & \cellcolor{yellow!50} & \cellcolor{green!30} & \cellcolor{green!30} & \cellcolor{yellow!50} & \cellcolor{yellow!50} \\
    & \cellcolor{yellow!50}\large{maple} & \cellcolor{green!30}\large{oak} & \cellcolor{green!30}\large{palm} & \cellcolor{yellow!50}\large{pine} & \cellcolor{yellow!50}\large{willow} \\
    & \cellcolor{yellow!50} & \cellcolor{green!30} & \cellcolor{green!30} & \cellcolor{yellow!50} & \cellcolor{yellow!50} \\
    
    \cellcolor{gray!20} & \cellcolor{yellow!50} & \cellcolor{yellow!50} & \cellcolor{green!30} & \cellcolor{yellow!50} & \cellcolor{green!30} \\
    \cellcolor{gray!20} & \cellcolor{yellow!50}\large{bicycle} & \cellcolor{yellow!50}\large{bus} & \cellcolor{green!30}\large{motorcycle} & \cellcolor{yellow!50}\large{pickup truck} & \cellcolor{green!30}\large{train} \\
    \multirow{-3}{*}{\cellcolor{gray!20}\large{vehicles $1$}} & \cellcolor{yellow!50} & \cellcolor{yellow!50} & \cellcolor{green!30} & \cellcolor{yellow!50} & \cellcolor{green!30} \\
    
    \multirow{3}{*}{\large{vehicles $2$}} & \cellcolor{green!30} & \cellcolor{yellow!50} & \cellcolor{yellow!50} & \cellcolor{green!30} & \cellcolor{yellow!50} \\
    & \cellcolor{green!30}\large{lawn-mower} & \cellcolor{yellow!50}\large{rocket} & \cellcolor{yellow!50}\large{streetcar} & \cellcolor{green!30}\large{tank} & \cellcolor{yellow!50}\large{tractor} \\
    & \cellcolor{green!30} & \cellcolor{yellow!50} & \cellcolor{yellow!50} & \cellcolor{green!30} & \cellcolor{yellow!50} \\
    \bottomrule
    \end{tabular}%
    }
    \vskip -0.1in
\end{table}

\subsection{CIFAR-based Benchmark}
The first benchmark dataset for the unknown detection task is based on CIFAR-100, a relatively small-scale dataset with low resolution images. As shown in Table~\ref{table:benchmark-setting}, CIFAR-100 has been frequently used for evaluations on all MD and OoDD tasks. In this work, CIFAR-100 is used to construct the in-distribution and the near-OoD dataset; the far-OoD datasets consists of four different ones: resized LSUN, SVHN, the Describable Textures Dataset (DTD)~\citep{cimpoi2014describing}, and Gaussian noise. Note that some works utilize the 80 million Tiny Images dataset~\citep{torralba2008tinyimages} if their methods require an external dataset, e.g., OE. However, the creators withdrew the dataset and requested that it not be used owing to some offensive and prejudicial contents which arose during the process of automatically collecting the images~\citep{birhane2021pyrrhic}. Hence, we use Tiny ImageNet~\footnote{\href{http://cs231n.stanford.edu/tiny-imagenet-200.zip}{http://cs231n.stanford.edu/tiny-imagenet-200.zip}}, a small-scale version of ImageNet, instead as the external dataset when necessary. 

\textbf{In-distribution \& Near-OoD datasets.}
CIFAR-100 is a dataset for 100-class classification, and its classes have a hierarchical structure. There are 20 superclasses with each having five subclasses, as shown in Table~\ref{table:cifar-split}. For example, \texttt{orchids}, \texttt{poppies}, \texttt{rose}, \texttt{sunflowers}, and \texttt{tulips} classes in CIFAR-100 belong to the superclass \texttt{flowers}. Based on this hierarchical structure, we can easily construct the in-distribution and near-OoD datasets, as it can be assumed that classes belonging to a specific superclass share some high-level semantics. Specifically, for each superclass, two classes are randomly selected as the in-distribution dataset, while the remaining three classes are considered as the near-OoD ones. As a result, 40 classes form the in-distribution dataset, termed \emph{CIFAR-40}, and 60 classes constitute the near-OoD dataset, referred to as \emph{CIFAR-60}, as shown in Table~\ref{table:cifar-split}.

\textbf{Far-OoD datasets.}
For the far-OoD datasets, we use datasets commonly employed as OoD datasets in the literature: resized LSUN, SVHN, DTD, and Gaussian Noise. To ensure that the far-OoD is semantically far from the in-distribution, we compose the far-OoD classes such that they do not share semantic concepts with in-distribution classes.

\begin{figure}[!t]
\vskip 0.1in
    \centering
    \includegraphics[width=0.6\textwidth]{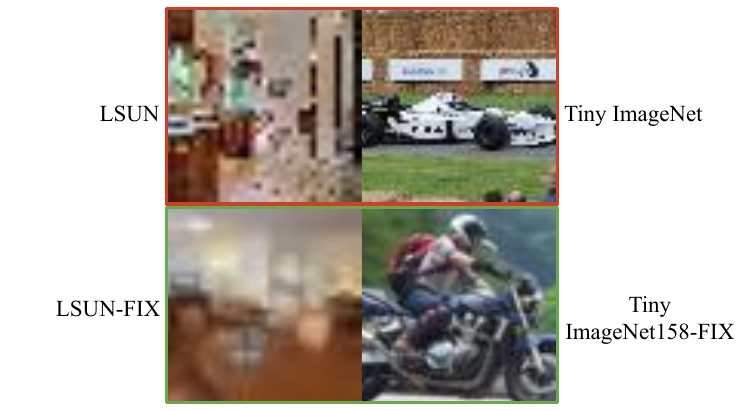}
    \caption{Examples of original images (red box; top row) and reconstructed images (green box; bottom row). The original images have artificial noise, leading to easy detection without considering the semantic information.}
    \label{fig:lsun-tiny}
\vskip -0.1in
\end{figure}

Resized LSUN with ten scene categories has been widely used in reference to the OoDD task~\citep{liang2017enhancing, lee2018simple}. However, as presented in the top row of Figure~\ref{fig:lsun-tiny}, images in resized LSUN contain artificial noise caused by inappropriate image processing, causing the OoDD performance to be overestimated significantly. \cite{tack2020csi} reported this problem by showing that resized LSUN is easily detectable with a simple smoothness score based on the total variation distance. Specifically, they defined total variation (TV) as the sum of $L_2$ distances between a given pixel and the surrounding pixels, summed over all pixels in an image. The smoothness score is then calculated as the difference between the average total variation over all images in a dataset and the total variation of each individual image. Hence, we reconstruct the dataset, referred to as LSUN-FIX, by employing a fixed resize operation following a procedure in~\cite{tack2020csi}. Samples from LSUN-FIX are shown in the bottom row of Figure~\ref{fig:lsun-tiny}.

We also empirically observed that the detection performance on LSUN reported in the literature is greatly overestimated, as shown in Section~\ref{sec:section4}. For example, in the literature, the Mahalanobis detector~\citep{lee2018simple}, a typical approach for OoDD, scored $98.2\%$ as the area under the receiver operating characteristic (AUROC) at the experimental setting of CIFAR-100 vs. LSUN. However, it showed $51.76\%$ AUROC in our experimental settings of detecting LSUN-FIX against CIFAR-40, indicating that the model virtually undertakes random guessing for the OoDD problem~\footnote{We excluded the Mahalanobis detector from the comparison based on this finding.}.

\begin{figure*}[!t]
    \centering
    \subfigure[CIFAR-100 subtree]{
        \includegraphics[width=0.3\columnwidth]{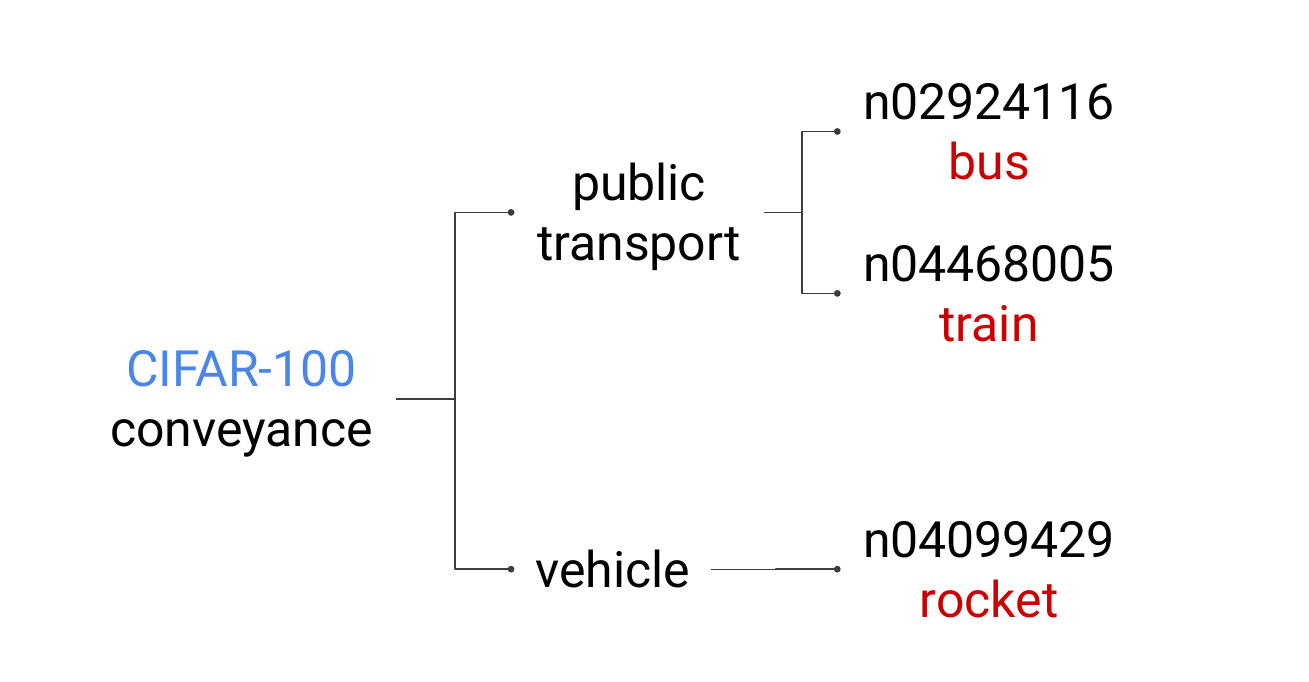}
        \label{fig:cifar-sub-a}
    }%
    \subfigure[Tiny ImageNet subtree]{
        \includegraphics[width=0.6\columnwidth]{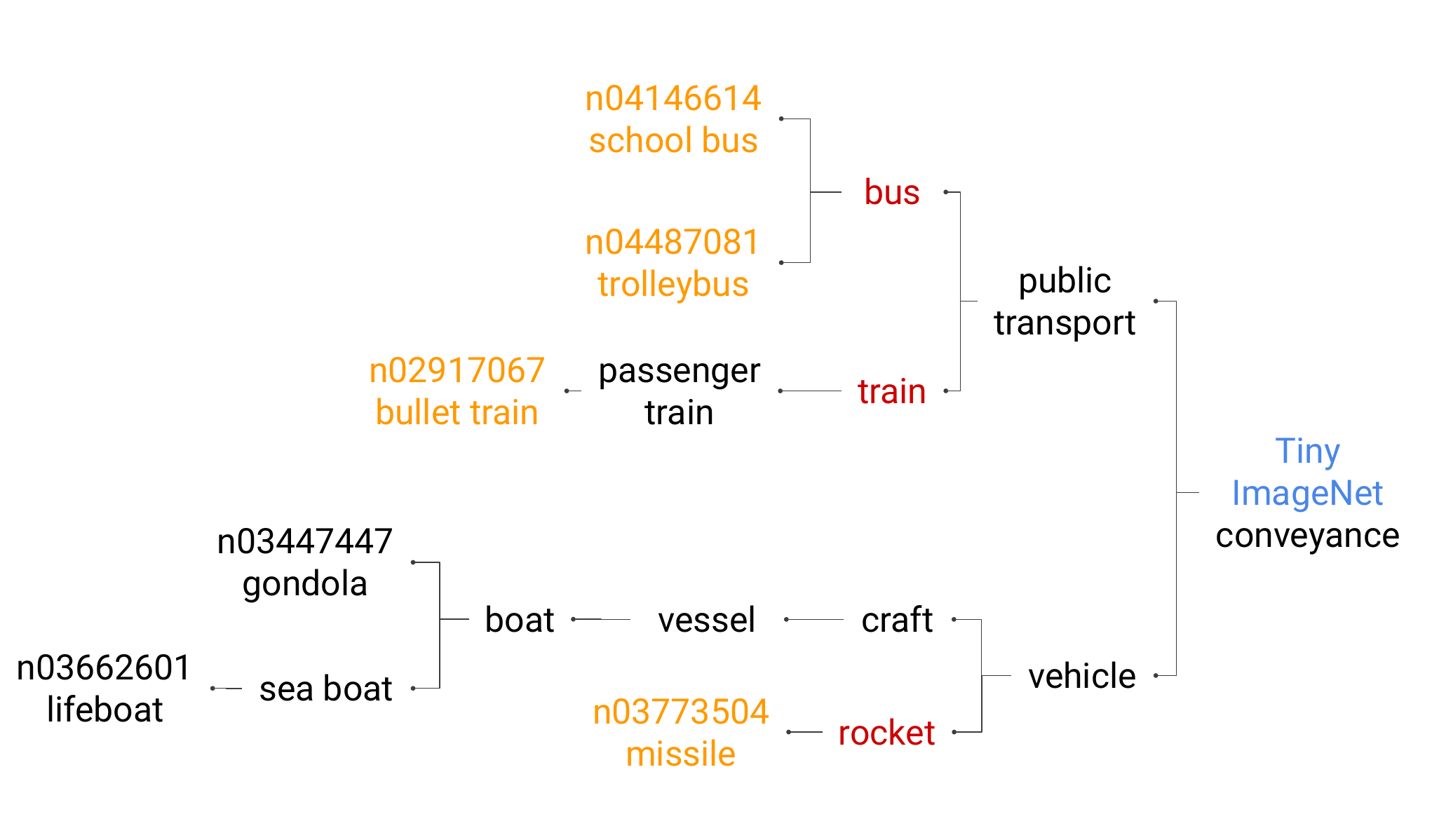}
        \label{fig:cifar-sub-b}
    }
    \caption{Part of (a) the CIFAR-100 subtree and (b) the Tiny ImageNet subtree. The nodes with \texttt{wnid} are classes of each dataset. The orange nodes are removed from Tiny ImageNet because they are descendant nodes of the CIFAR-100 classes.}
    \label{fig:cifar-sub}
\vskip -0.1in
\end{figure*}

It is important to note that classes in the far-OoD datasets should not be similar to those in the in-distribution dataset \emph{CIFAR-40}. To examine the similarity between classes in these datasets, we utilize the WordNet tree, which shows a hierarchical structure of concepts by encoding a super-subordinate relationship of words.
First, the subtree having nodes corresponding to the CIFAR-100 classes is extracted from the overall WordNet tree. One part of this subtree is illustrated in Figure~\ref{fig:cifar-sub-a}. Then, a particular class in the far-OoD datasets is removed if its class name matches at least one node in the subtree. Obviously, the classes in the SVHN, DTD, and Gaussian noise datasets do not need to be examined. For LSUN-FIX, it is confirmed that all classes do not overlap with those in CIFAR-100, therefore we did not remove any classes from LSUN-FIX. Hence, it is guaranteed that our far-OoD datasets, SVHN, DTD, Gaussian noise, and LSUN-FIX, do not semantically overlap with the in-distribution dataset \emph{CIFAR-40}.

\textbf{External dataset.}
Tiny ImageNet consists of 200 classes with low resolution images from the original ImageNet dataset~\citep{le2015tiny}. It also has downsampled images with artificial noise, leading to the same problem found in the resized LSUN dataset, as shown in Figure~\ref{fig:lsun-tiny}. Thus, we reconstructed the Tiny ImageNet dataset by resizing the corresponding images of ImageNet with a fixed resize operation as used to construct LSUN-FIX~\footnote{Hereafter, the term "Tiny ImageNet-FIX" refers to the reconstructed Tiny ImageNet dataset.}.

As mentioned earlier, the external dataset should not overlap with the in-distribution or the near-OoD dataset. Images of CIFAR-100 and Tiny ImageNet were collected based on certain words in WordNet. Each word in WordNet has its own WordNetID (\texttt{wnid}), meaning that \texttt{wnid} can be used to examine the overlap between classes in CIFAR-100 and Tiny ImageNet on the WordNet tree. However, the classes in CIFAR-100 do not have explicit \texttt{wnid}, while those in Tiny ImageNet do. Accordingly, we manually assigned the corresponding \texttt{wnid} to the CIFAR-100 classes. For example, the CIFAR-100 classes \texttt{camel} and \texttt{mouse} are assigned \texttt{wnid} \texttt{n02437136} and \texttt{n02330245}, respectively.

The procedure used to examine the similarity between CIFAR-100 classes and Tiny ImageNet classes is conducted on a subtree of the WordNet tree. Here, the subtree consists of nodes corresponding to the classes in Tiny ImageNet, whereas the subtree used to construct the far-OoD datasets consists of those in CIFAR-100. A particular class in Tiny ImageNet is removed if its \texttt{wnid} is identical to a specific \texttt{wnid} of CIFAR-100 classes or is a descendant of nodes corresponding to CIFAR-100 classes. Figure~\ref{fig:cifar-sub-b} presents part of this subtree. The leaf nodes with \texttt{wnid} and the red nodes indicate classes in Tiny ImageNet and CIFAR-100, respectively. Orange nodes such as \texttt{bullet train} or \texttt{school bus} have CIFAR-100 classes  (i.e., \texttt{train} and \texttt{bus}) as their ancestors, and they can be regarded as classes semantically overlapping the CIFAR-100 classes. According to this procedure, we removed such classes (i.e., orange nodes) from the Tiny ImageNet dataset. As a result, 42 classes in total were removed from Tiny ImageNet; therefore, the remaining 158 classes comprise Tiny ImageNet158-FIX, which is the external dataset of our CIFAR-based benchmark. 

\subsection{ImageNet-based Benchmark}
The well-known ImageNet is the most popular large-scale dataset typically used for image classification. 
To evaluate the unknown detection performance with a large-scale dataset, we also developed an ImageNet-based benchmark following similar procedures employed for our CIFAR-based benchmark. The WordNet tree fits perfectly when used to determine the semantic distance between ImageNet classes, as all classes of ImageNet come from WordNet~\citep{deng2009imagenet}. To construct our ImageNet-based benchmark, ImageNet classes were divided into three categories based on a hierarchical structure of the WordNet tree: the in-distribution, the near-OoD, and the external datasets. As the far-OoD datasets, we considered Food-101~\citep{bossard2014food}, Caltech-256~\citep{griffin2007caltech}, and Places-365~\citep{zhou2018places} after removing classes that overlapped with the ImageNet classes.

\begin{figure*}[!t]
\vskip 0.1in
    \subfigure[near-OoD/external classes]{
        \includegraphics[width=0.47\textwidth]{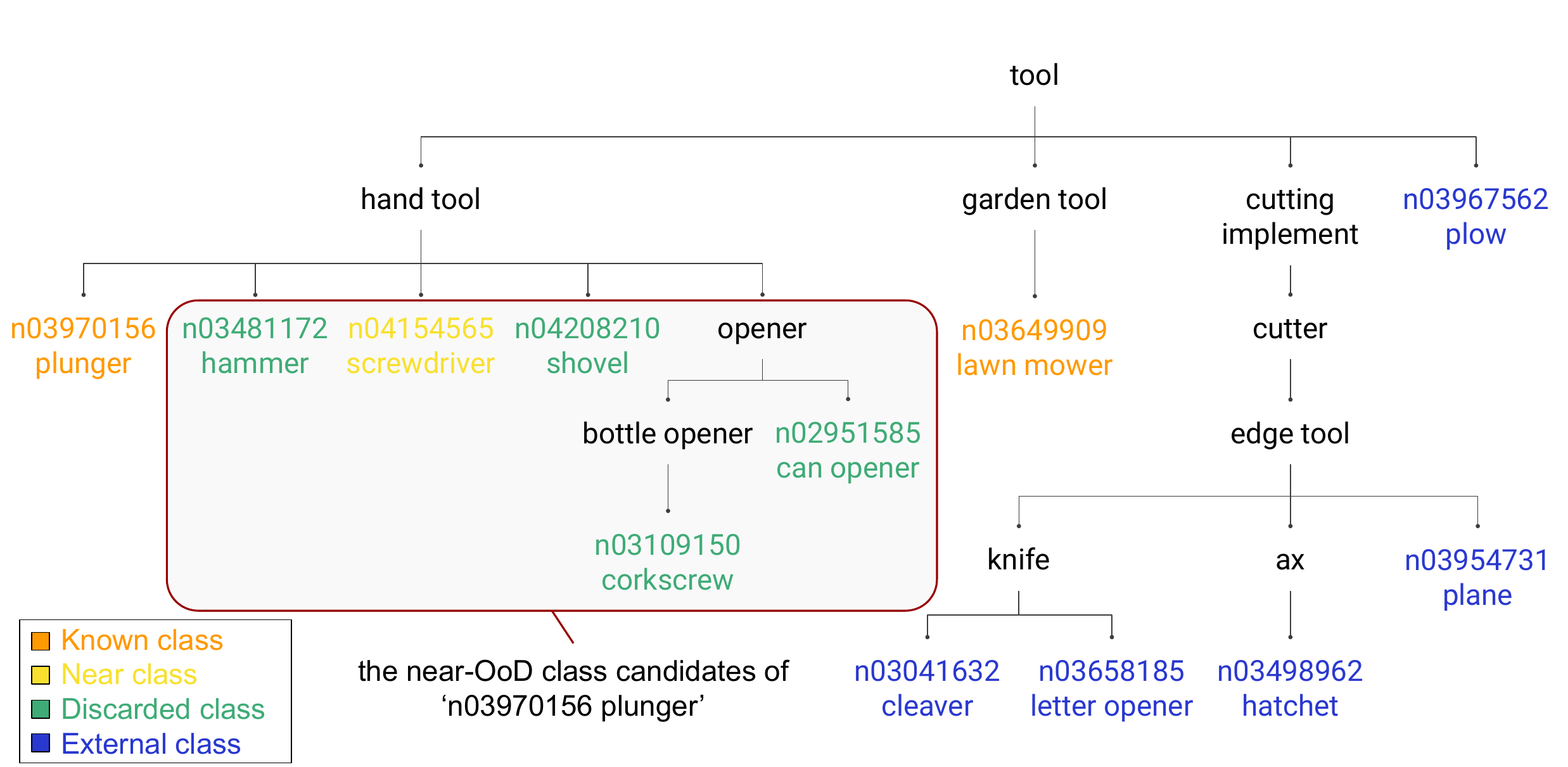}
        \label{fig:imagenet-sub-a}
    }
    \subfigure[far-OoD classes]{
        \includegraphics[width=0.47\textwidth]{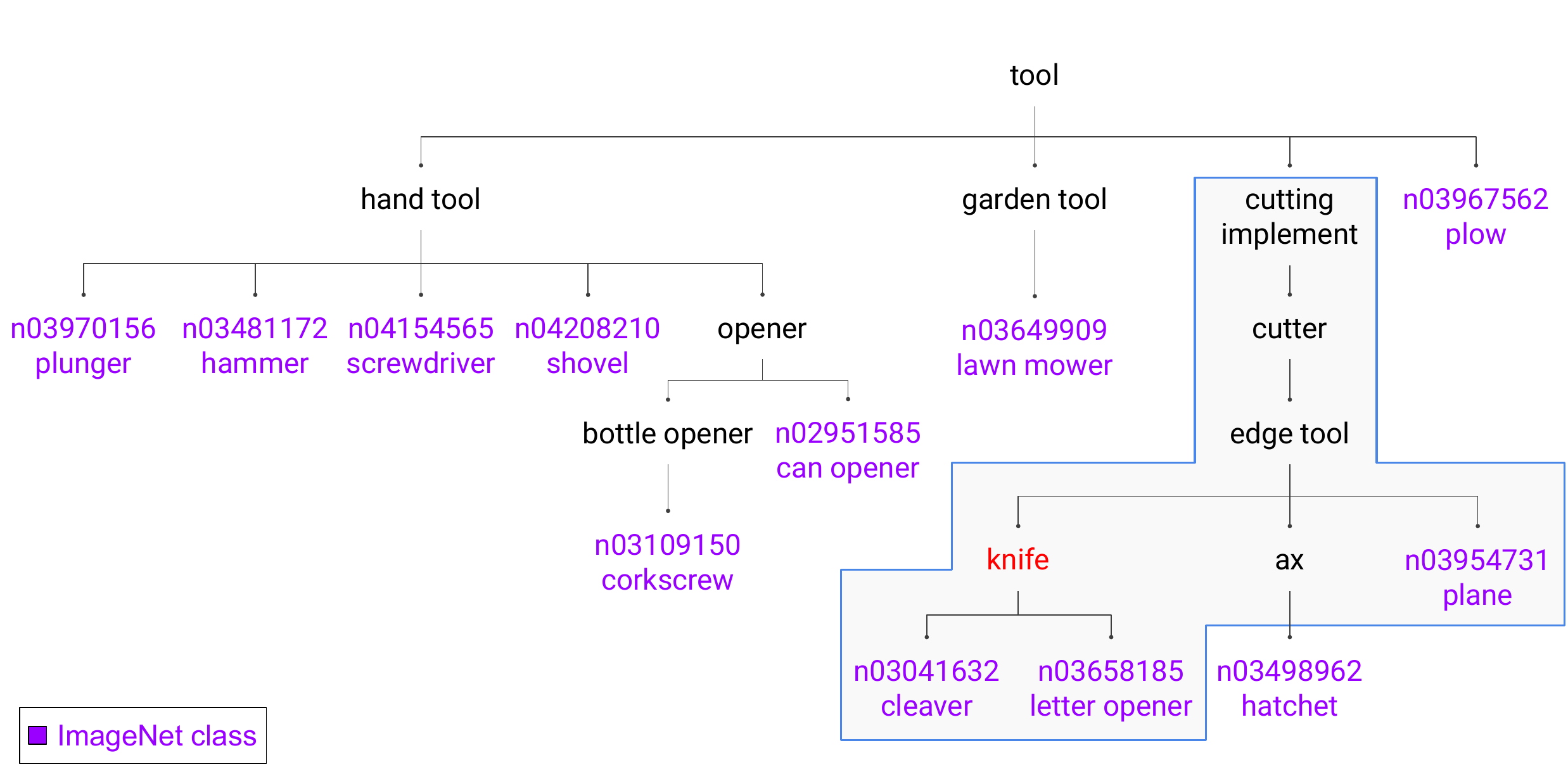}
        \label{fig:imagenet-sub-b}
    }
    \caption{Construction process of the near-/far-OoD, and the external datasets in ImageNet-based benchmark: (a) shows how the near-OoD classes and the external classes are determined. The classes in the red box are near-OoD class candidates with the in-distribution class \texttt{plunger}, and (b) explains how the far-OoD classes are examined. The blue box contains nodes to be examined to determine if a class having the keyword \texttt{knife} should be removed from the far-OoD datasets.}
    \label{fig:imagenet-sub}
\vskip 0.1in
\end{figure*}

\textbf{In-distribution dataset.}
We constructed the in-distribution dataset \emph{ImageNet-200} with 200 classes identical to those in Tiny ImageNet. Note that the images in this benchmark are not downsampled to preserve the original image resolutions. Specifically, all ImageNet training images from the 200 classes were used to construct the training and validation datasets for our ImageNet-based benchmark.

\textbf{Near-OoD \& External datasets.}
To categorize ImageNet classes according to the degree of semantic-level similarity, we utilize the WordNet tree, as in the procedure to examine the far-OoD and external classes in the CIFAR-based benchmark. The subtree of WordNet consists of nodes corresponding to ImageNet 1000 classes. Figure~\ref{fig:imagenet-sub} shows a part of the subtree. We regard classes that are descendants of the parent nodes of in-distribution classes (i.e., ImageNet-200) as near-OoD class candidates. For example, in Figure~\ref{fig:imagenet-sub-a}, \texttt{can opener} is one of the near-OoD class candidates because it is a descendant node of \texttt{hand tool} which is the parent node of the in-distribution class \texttt{plunger}. On the other hand, \texttt{garden tool} which is the parent node of the in-distribution class \texttt{lawn mower}, does not have any descendants. Accordingly, there are no near-OoD class candidates under \texttt{garden tool}.

Only 200 classes are randomly selected among all near-OoD candidates, while the remaining candidates are discarded. The external dataset consists of the remaining 394 ImageNet classes, excluding 200 in-distribution classes, 200 near-OoD classes, and the discarded classes. Consequently, the selected 200 classes formed a near-OoD dataset referred to as \emph{Near ImageNet-200}, and the remaining 394 classes formed the external dataset \emph{External ImageNet-394}. Note that \emph{Near ImageNet-200} consists of ImageNet validation images corresponding to 200 near-OoD classes, whereas \emph{External ImageNet-394} is constructed using ImageNet training images from 394 external-OoD classes.

\textbf{Far-OoD dataset.}
For the far-OoD datasets, we adopt other high-resolution datasets, Food-101, Caltech-256 and Places-365. To make the far-OoD datasets clearly semantically far from the in-distribution dataset, we eliminate classes similar to the in-distribution classes from them. The keywords in the names of the far-OoD classes are used as the starting point for the similarity examination process.

We utilize the same subtree used to examine the near-OoD classes in the ImageNet-based benchmark. First, we inspect if there are any nodes matching a keyword in the names of the far-OoD classes. A far-OoD class will be removed if \romannumeral 1 $)$ the matching node is one of the ImageNet classes, \romannumeral 2 $)$ sibling nodes of the matching node contain at least one of the ImageNet classes, or \romannumeral 3 $)$ ancestor or descendant nodes of the matching node have at least one of the ImageNet classes.
Figure~\ref{fig:imagenet-sub-b} gives an example of the examination process when the name of a certain far-OoD class includes \texttt{knife}. The subtree has a node including \texttt{knife} in its name. Therefore, we check all of the sibling, descendant, and ancestor nodes (see the blue box in the figure). Far-OoD classes having \texttt{knife} in their names will be removed because they meet the aforementioned removal criteria \romannumeral 2 $)$ or \romannumeral 3 $)$.

\begin{table}[!t]
    \centering
    \caption{Examples of grammatical variations of the keywords in each far-OoD dataset. The left side of the arrow is the original class name and the right side is its possible variation.}
    \label{table:imagenet-name}
    \resizebox{0.7\textwidth}{!}{%
    \begin{tabular}{l|ccc}
    \toprule
    \textbf{\large{Variations}} & \textbf{\large{Food101}} & \textbf{\large{Caltech256}} & \textbf{\large{Places365}} \\ \midrule
    
    \multicolumn{1}{l|}{\begin{tabular}[l]{@{}c@{}} \large{Plural} \end{tabular}} &
    \multicolumn{1}{c}{\begin{tabular}[c]{@{}c@{}} \large{cup cakes $\rightarrow$ cup cake} \\ \large{donuts $\rightarrow$ donut} \\ \large{oysters $\rightarrow$ oyster} \end{tabular}} &
    \multicolumn{1}{c}{\begin{tabular}[c]{@{}c@{}} \large{billiards} $\rightarrow$ \large{billiard} \\ \large{chopsticks $\rightarrow$ chopstick} \\ \large{head phones $\rightarrow$ head phone} \end{tabular}} &
    \multicolumn{1}{c}{\begin{tabular}[c]{@{}c@{}} \large{badlands $\rightarrow$ badland} \\ \large{artists loft $\rightarrow$ artist loft} \\ \large{butchers shop $\rightarrow$ butcher shop} \\ \large{childs room $\rightarrow$ child room} \end{tabular}}  \\ \midrule
    
    \multicolumn{1}{l|}{\begin{tabular}[l]{@{}c@{}} \large{Spacing} \end{tabular}} &
    \multicolumn{1}{c}{\begin{tabular}[c]{@{}c@{}} \large{cup cakes $\rightarrow$ cupcakes} \\ \large{hot dog $\rightarrow$ hotdog} \end{tabular}} &
    \multicolumn{1}{c}{\begin{tabular}[c]{@{}c@{}} \large{chess board $\rightarrow$ chessboard} \\ \large{dumb bell $\rightarrow$ dumbbell} \\ \large{head phones $\rightarrow$ headphones}\end{tabular}} &
    \multicolumn{1}{c}{\begin{tabular}[c]{@{}c@{}} \large{barndoor $\rightarrow$ barn door} \\ \large{fishpond $\rightarrow$ fish pond} \\ \large{shopfront $\rightarrow$ shop front} \end{tabular}}
    \\ \midrule
    
    \multicolumn{1}{l|}{\begin{tabular}[l]{@{}c@{}} \large{Abbreviation} \end{tabular}} &
    \multicolumn{1}{c}{\begin{tabular}[c]{@{}c@{}} - \end{tabular}} &
    \multicolumn{1}{c}{\begin{tabular}[c]{@{}c@{}} \large{chimp $\rightarrow$ chimpanzee} \end{tabular}} &
    \multicolumn{1}{c}{\begin{tabular}[c]{@{}c@{}} - \end{tabular}}
    \\ \midrule
    
    \multicolumn{1}{l|}{\begin{tabular}[l]{@{}c@{}} \large{Apostrophe} \end{tabular}} &
    \multicolumn{1}{c}{\begin{tabular}[c]{@{}c@{}} - \end{tabular}} &
    \multicolumn{1}{c}{\begin{tabular}[c]{@{}c@{}} - \end{tabular}} &
    \multicolumn{1}{c}{\begin{tabular}[c]{@{}c@{}} \large{artist loft $\rightarrow$ artist's loft} \\ \large{butchers shop $\rightarrow$ butcher's shop} \\ \large{child room $\rightarrow$ child's room} \end{tabular}} \\
    
    \bottomrule
    \end{tabular}%
    }
    \vskip -0.1in
\end{table}

Each dataset has its own naming rule and we, therefore, consider grammatical variations of the keywords, such as plurals, abbreviations, and different spacing, among others. For example, for \texttt{cup cakes} in Food-101, \texttt{cup}, \texttt{cups}, \texttt{cake}, \texttt{cakes}, \texttt{cupcake}, and \texttt{cupcakes} are considered keywords. In a case such as \texttt{chimp} in Caltech-256, we check images in the class to clarify whether it is a short form of \texttt{chimpanzee}, and consider \texttt{chimpanzee} as a keyword for that class. Table~\ref{table:imagenet-name} provides several examples of grammatical variations of the types considered in this work.

The Food-101, Caltech-256, and Places-365 classes have 32, 45, and 82 remaining classes after the examining processes, as described above, and they were renamed Food-32, Caltech-45, and Places-82, respectively. Consequently, the far-OoD datasets are constituted by images from the remaining classes of Food-101, Caltech-256, and Places-365.

\section{Experimental Results} \label{sec:section4}
In this section, we present the comparison methods, experimental settings and evaluation metrics for the proposed unknown detection task. Each following subsection provides the unknown detection performances of the comparison methods on the proposed CIFAR- and ImageNet-based benchmarks.

\textbf{Comparison methods.}
We considered a total of ten methods including methods belonging to categories relevant to the unknown detection task: MD and OoDD. As a performance baseline, we employed the approach of using the vanilla maximum class probability (MCP)~\citep{hendrycks2016baseline} produced by the baseline model as a confidence score, denoted as ``MCP".

Among the existing methods developed for MD, we chose two methods: CRL~\citep{moon2020confidence} and EDL~\citep{sensoy2018edl}. CRL utilizes a ranking loss based on the frequency of correct predictions for each sample to better estimate confidence scores. As a result, a model trained with CRL produces high confidence scores on correctly predicted samples. EDL treats the softmax output as a categorical distribution by introducing the Dirichlet density so that the predictions can be interpreted as a distribution over possible softmax outputs.

For the OoDD methods, we chose five popular methods: ODIN~\citep{liang2017enhancing}, Outlier Exposure (OE)~\citep{hendrycks2018deep}, Energy-based OoDD (Energy)~\citep{liu2020energy}, ReAct~\citep{sun2021react}, and OpenMax~\citep{bendale2016towards}. ODIN broadens a gap in the confidence values between in-distribution and OoD inputs through temperature scaling and by adding small perturbations to the input samples. OE suggested a training method that leverages an external dataset consisting of outliers. By explicitly exposing a model to such outliers, OE substantially improves the OoDD performance. Energy restricts the OoD samples to have higher energy values than in-distribution samples. ReAct is a post-hoc method that truncates the activation values. It is based on the finding that OoD samples generally produce activation patterns that differ substantially from those of ID data. OpenMax utilizes extreme value theory to estimate the distribution of inputs that are not likely to be the in-distribution with a Weibull distribution.

We also evaluated several previous methods which regularize a model to have a high quality of confidence estimates: AugMix~\citep{hendrycks2019augmix}, Monte Carlo dropout (MCdropout)~\citep{gal2016dropout}, and Deep Ensemble. AugMix suggested a data augmentation technique that mixes several randomly chosen data augmentations. It is focused on improving both the robustness and uncertainty estimates of the predictions of a model under a distribution shift, such as corrupted inputs. MCdropout enables estimations of the posterior distribution by sampling several stochastic predictions using a dropout technique during the test phase. For MCdropout, confidence values are estimated by averaging $50$ stochastic predictions~\citep{kendall2017aleatoric}. The ResNet architecture for MCdropout used in our experiments stems from \cite{zhang2019confidence}. Specifically, dropout layers with a dropout rate of $p=0.2$ are applied to all convolutional layers, and a dropout layer with $p=0.1$ is added before the last fully connected layer. For Deep Ensemble, confidence values are estimated by averaging predictions from five baseline models~\footnote{Both MCdropout and Deep Ensemble offer various options for computing confidence scores, such as MCP or entropy of the average predictions, or variance over predictions. We empirically found that MCP and entropy provide similar performance for unknown detection, while variance was found to be suboptimal. Hence, we report the results obtained using MCP for both MCdropout and Deep Ensemble.}.

For all comparison methods, we employed the confidence scoring schemes specified in the original literature because they are expected to offer a clear separation between known and unknown samples if each method possesses the desired property of reliable confidence estimates.

\begin{table}[!t]
    \centering
    \caption{Summary of the search range and selected hyperparameters for each comparison method.}
    \label{table:hyperparameter}
    \resizebox{\textwidth}{!}{%
    \begin{tabular}{l|ccc}
    \toprule
    \textbf{Method} & \textbf{Hyperparameter} & \textbf{Search Range} & \begin{tabular}[c]{@{}c@{}} \textbf{Selected Value} \\ \textbf{CIFAR- / ImageNet-based} \end{tabular} \\ \midrule
    CRL & Loss weight $\lambda$ & - & $1.0$ \\ \midrule
    \multirow{2}{*}{EDL} & Objective type & MSE, digamma, log, cross-entropy & log \\
    & Output activation & ReLU, exponential, softplus & exponential \\ \midrule
    \multirow{2}{*}{ODIN} & Temperature $\tau$ & $1, 10, 100, 1000$ & $10$ / $10$ \\
    & Perturbation magnitude $\epsilon$ & $0, 0.0005, 0.001, 0.0014, 0.002, 0.0024, 0.005, 0.01, 0.05, 0.1, 0.2$ & $0.002$ / $0$ \\ \midrule
    OE & Loss weight $\lambda$ & - & $0.5$ \\ \midrule
    \multirow{3}{*}{Energy} & Margin for ID $m_{\text{in}}$ & $-3, -5, -7$ & $-3$ / $-7$ \\ 
    & Margin for OoD $m_{\text{out}}$ & $-15, -19, -23, -27$ & $-27$ / $-15$ \\
    & Finetuning epoch & - & $10$ \\ \midrule
    ReAct & Logit threshold $c$ & $1,2,...,10$ & $7$ / $2$ \\ \midrule
    \multirow{3}{*}{OpenMax} & The number of classes $\alpha$ & $1, 5, 10, 20, 30, 40$ & $1$ \\
    & \begin{tabular}[c]{@{}c@{}} The number of samples $\eta$ \end{tabular} & $2, 5, 10, 20, 30, 40$ & $2$ \\
    & Distance weight & - & $0.0005$ \\ \midrule
    \multirow{4}{*}{AugMix} & Mixture width $k$ & - & $3$ \\
    & Dirichlet parameter $\alpha$ & - & $1.0$ \\ 
    & Augmentation severity & - & $3$ \\
    & JSD consistency loss & True, False & False \\ \midrule
    \bottomrule
    \end{tabular}%
    }
    \vskip -0.1in
\end{table}

\textbf{Hyperparameter search.}
We adopted the reported hyperparameters from the original literature of each method, provided that the corresponding literature revealed that the hyperparameters used were not significantly affected by experimental settings such as training datasets or model architectures. Alternatively, in cases where the hyperparameters were found to be significantly affected by the experimental settings, we performed a search to determine the optimal values based on the search range provided in the literature. For MD methods, the validation metric was the classification accuracy on the in-distribution validation set. For OoDD methods, we evaluated the hyperparameters based on the OoDD performance in terms of AUROC on our validation sets (refer to Table~\ref{table:benchmark-summary}), as they were designed for this purpose. Note that the external dataset does not overlap with the near- and far-OoD datasets, which ensures that a model cannot directly access the test datasets (i.e., the constructed OoD datasets). Table~\ref{table:hyperparameter} summarizes the search ranges and the selected hyperparameters of each comparison method on each benchmark.

\textbf{Experimental settings.} We evaluate all comparison methods under the same experimental settings on each benchmark. For the CIFAR-based benchmark, we employed the ResNet18~\footnote{\href{https://github.com/junyuseu/pytorch-cifar-models/blob/master/models/resnet\_cifar.py}{https://github.com/junyuseu/pytorch-cifar-models/blob/master/models/resnet\_cifar.py}}~\citep{he2016identity} architecture. All models were trained using SGD with a momentum of $0.9$, an initial learning rate of $0.1$, and a weight decay of $0.0005$ for $200$ epochs with a mini-batch size of $128$. The learning rate was reduced by a factor of $10$ at $120$ epochs and $160$ epochs. Also, we employed a standard data augmentation scheme, i.e., random horizontal flip and $32\times32$ random cropping after padding with $4$ pixels on each side.

For the ImageNet-based benchmark, we adopted standard ImageNet training settings with the ResNet152 architecture~\footnote{\href{https://github.com/pytorch/examples/tree/master/imagenet}{https://github.com/pytorch/examples/tree/master/imagenet}}. Every model was trained using SGD with a momentum of $0.9$, an initial learning rate of $0.1$, and a weight decay of $0.0001$ for $90$ epochs with a mini-batch size of $256$. The learning rate was reduced by a factor of $10$ at $30$ epochs and $60$ epochs. The data augmentation scheme consisted of $224\times224$ random resized cropping and random horizontal flip.

OE was trained from scratch and the mini-batch size of the external dataset (i.e., Tiny ImageNet158-FIX or ImageNet-394 for CIFAR- and ImageNet-based benchmark, respectively) was set to twice that of the in-distribution dataset following the original training recipe~\citep{hendrycks2018deep}. Energy finetunes the pretrained model using both in-distribution and external datasets for better OoDD performance. Following \cite{liu2020energy}, we finetuned the pretrained model during $10$ epochs, and the mini-batch size of the external dataset was set to twice the size of the in-distribution mini-batch size.

\textbf{Evaluation metrics.} 
The most commonly used performance metric for MD and OoDD is the area under the receiver operating characteristic curve (AUROC). Because these tasks are essentially intended to distinguish between correct predictions (or in-distribution samples) and incorrect ones (or OoD samples), measuring the ranking performance of confidence values can be a natural choice. However, evaluating unknown detection performances using AUROC could be problematic. As \cite{ashukha2020pitfalls} pointed out, AUROC cannot be used directly to compare MD performances across different methods, as each model has its own correct and incorrect predictions. In other words, each model induces an individual classification problem with its own positives (i.e., correct predictions) and negatives (i.e., incorrect predictions). Similar to the MD problem, the unknown detection task considers incorrect predictions as unknown samples; accordingly, AUROC is not suitable for precisely comparing unknown detection performances between different methods.

We thus adopt the area under the risk-coverage curve (AURC)~\citep{geifman2018bias} as a primary measure of the unknown detection performance. AURC measures the area under the curve of risk at the sample coverage. The sample coverage is defined as the ratio of samples whose confidence values exceed a certain confidence threshold. Let $\mathcal{D}=\{(\mathbf{x}_i, y_i)\}_{i=1}^{n}$ be a dataset consisting of $n$ labeled samples from a joint distribution over $\mathcal{X}\times\mathcal{Y}$, where $\mathcal{X}$ is the input space and $\mathcal{Y}=\{1,2,3, \ldots , K\}$ is the label set for the classes. Given a softmax classifier $f(\mathbf{x};\mathbf{w})$ with a model parameter set $\mathbf{w}$, we can obtain the predicted class probabilities $\mathbf{p}_i=P(y|\mathbf{x}_i, \mathbf{w})$ for each $\mathbf{x}_i$ and confidence score $\kappa(\mathbf{p}_i)$, e.g., the maximum class probability or entropy, associated with the prediction. Then, a subset of $\mathcal{D}$, $\mathcal{S}_{\theta}$, consisting of samples whose confidence scores $\kappa(\mathbf{p}_i)$ are above a predefined threshold $\theta$ can be constructed. Therefore, the sample coverage at the confidence threshold $\theta$, $c(\theta)$ is computed as

\begin{eqnarray}
\label{formula: coverage}
    && c(\theta) = \frac{|\mathcal{S}_{\theta}|}{|\mathcal{D}|}
    \nonumber
\end{eqnarray}
where $|\cdot|$ denotes the cardinality of a set.
The risk represents the error rate of the samples in $\mathcal{S}_{\theta}$. Therefore, the risk at $\theta$, $r(\theta)$, can be simply computed as 
\begin{eqnarray}
\label{formula: risk}
    && r(\theta) = \frac{
    \sum_{(\mathbf{x}_i,y_i) \in \mathcal{S}_{\theta}} \mathbbold{1}(\hat{y}_{i}\neq y_i)
    }
    {|\mathcal{S}_{\theta}|}
    \nonumber
\end{eqnarray}
where $\mathbbold{1}(\cdot)$ denotes an indicator function and $\hat{y}_{i}$ represents the predicted class, i.e., $\underset{y\in\mathcal{Y}}{\arg\max}P(y|\mathbf{x}_i, \mathbf{w})$. Note that OoD samples from both near- and far-OoD should be considered as errors in our unknown detection task since, by definition, their true classes do not belong to any of the in-distribution classes. Consequently, a low AURC value implies that $f(\mathbf{x};\mathbf{w})$ assigns higher confidence values to correctly predicted in-distribution samples as compared to incorrect predictions and OoD samples. In other words, a classifier with lower AURC would be more reliable in that it shows strong confidence only for what it knows. Note that for all comparison methods, we employed the confidence scoring scheme $\kappa(\mathbf{p}_i)$ specified in the original literature.

Although our primary goal is to examine the unknown detection performance of comparison targets based on the AURC, we also evaluate their MD and OoDD performances separately to gain useful insight into the behavior of each method. For example, it has yet to be revealed as to whether methods showing superior OoDD performance are effective with MD as well, and vice versa. To this end, we additionally consider other performance metrics that are commonly used for OoDD: AUROC and the false positive rate at the 95\% true positive rate (FPR-95\%-TPR). Note that the MD performance is measured by AURC following~\cite{geifman2018bias} and \cite{moon2020confidence}.

\begin{table*}[!t]
    \centering
    \caption{Comparison of unknown detection performance with other methods on the CIFAR-based benchmark. The means and standard deviations over five runs are reported. $\downarrow$ and $\uparrow$ indicate that lower and higher values are better, respectively. \textcolor{red}{\textbf{Red}} represents the best performance among all methods, whereas \textcolor{blue}{\textbf{blue}} represents the best performance of the other methods except those trained on the auxiliary dataset (i.e., Outlier Exposure and Energy). The AURC values are multiplied by $10^3$, and the other values are percentages.}
    \label{table:cifar-perf}
    \vskip 0.1in
    \resizebox{1.0\textwidth}{!}{%
    \begin{tabular}{l|c|c|c|c|cccc}
    \toprule
    
    \multicolumn{1}{c}{\textbf{\begin{tabular}[c]{@{}c@{}} \\ \end{tabular}}} &
    \multicolumn{1}{c|}{\textbf{\begin{tabular}[c]{@{}c@{}} \\ \end{tabular}}} &
    \multicolumn{1}{c|}{\textbf{\begin{tabular}[c]{@{}c@{}} Unknown \end{tabular}}} &
    \textbf{\begin{tabular}[c]{@{}c@{}} Miscls. \end{tabular}} &
    \textbf{\begin{tabular}[c]{@{}c@{}} Near-OoD \end{tabular}} &
    \multicolumn{4}{c}{\textbf{\begin{tabular}[c]{@{}c@{}} Far-OoD Detection \end{tabular}}} \\ [3pt]
    
    \multicolumn{1}{c}{\textbf{\begin{tabular}[c]{@{}c@{}}  \end{tabular}}} &
    \multicolumn{1}{c|}{\textbf{\begin{tabular}[c]{@{}c@{}} \end{tabular}}} &
    \textbf{\begin{tabular}[c]{@{}c@{}} Detection \end{tabular}} &
    \textbf{\begin{tabular}[c]{@{}c@{}} Detection \end{tabular}} &
    \textbf{\begin{tabular}[c]{@{}c@{}} Detection \end{tabular}} &
    \textbf{\begin{tabular}[c]{@{}c@{}} SVHN \end{tabular}} &
    \textbf{\begin{tabular}[c]{@{}c@{}} LSUN-FIX \end{tabular}} &
    \textbf{\begin{tabular}[c]{@{}c@{}} DTD \end{tabular}} &
    \textbf{\begin{tabular}[c]{@{}c@{}} Gaussian Noise \end{tabular}} \\ \hline
    
    \textbf{Method} & \textbf{ACC($\uparrow$)} & \multicolumn{2}{c|}{\textbf{AURC($\downarrow$)}} & \multicolumn{5}{c}{\textbf{\begin{tabular}[c]{@{}c@{}} FPR-95$\%$-TPR($\downarrow$) \\ AUROC($\uparrow$) \end{tabular}}} \\ \hline
    
    MCP & 74.43$\pm$0.51 & 773.15$\pm$12.77 & 75.25$\pm$2.68 &
    \begin{tabular}[c]{@{}c@{}} 86.50$\pm$0.54 \\ 72.16$\pm$0.36 \end{tabular} &
    \begin{tabular}[c]{@{}c@{}} 85.92$\pm$1.80 \\ 74.91$\pm$1.30 \end{tabular} &
    \begin{tabular}[c]{@{}c@{}} 87.02$\pm$0.32 \\ 70.50$\pm$0.96 \end{tabular} &
    \begin{tabular}[c]{@{}c@{}} 87.90$\pm$0.93 \\ 70.69$\pm$1.29 \end{tabular} &
    \begin{tabular}[c]{@{}c@{}} 84.61$\pm$21.31 \\ 72.10$\pm$13.98 \end{tabular} \\ [12pt]
    
    CRL & 75.15$\pm$0.42 & 770.17$\pm$22.19 & 68.53$\pm$2.83 &
    \begin{tabular}[c]{@{}c@{}} 85.69$\pm$1.33 \\ 72.95$\pm$0.51 \end{tabular} &
    \begin{tabular}[c]{@{}c@{}} 82.74$\pm$2.55 \\ 78.38$\pm$1.66 \end{tabular} &
    \begin{tabular}[c]{@{}c@{}} 87.93$\pm$0.81 \\ 69.66$\pm$0.30 \end{tabular} &
    \begin{tabular}[c]{@{}c@{}} 87.90$\pm$0.93 \\ 71.90$\pm$1.34 \end{tabular} &
    \begin{tabular}[c]{@{}c@{}} 91.39$\pm$8.19 \\ 69.11$\pm$15.53 \end{tabular} \\ [12pt]
    
    EDL & 74.40$\pm$0.26 & 763.67$\pm$6.95 & 73.39$\pm$1.39 &
    \begin{tabular}[c]{@{}c@{}} 86.23$\pm$0.50 \\ 72.72$\pm$0.29 \end{tabular} &
    \begin{tabular}[c]{@{}c@{}} 83.12$\pm$3.03 \\ 78.20$\pm$1.26 \end{tabular} &
    \begin{tabular}[c]{@{}c@{}} 86.81$\pm$0.79 \\ 71.32$\pm$0.21 \end{tabular} &
    \begin{tabular}[c]{@{}c@{}} 86.46$\pm$0.57 \\ 73.32$\pm$0.37 \end{tabular} &
    \begin{tabular}[c]{@{}c@{}} 96.86$\pm$4.03 \\ 72.05$\pm$5.02 \end{tabular} \\ [12pt]
    
    MCdropout & 76.00$\pm$0.48 & 791.99$\pm$9.22 & 66.16$\pm$1.61 &
    \begin{tabular}[c]{@{}c@{}} 86.71$\pm$0.63 \\ 72.37$\pm$0.11 \end{tabular} &
    \begin{tabular}[c]{@{}c@{}} 82.73$\pm$1.58 \\ 79.17$\pm$1.09 \end{tabular} &
    \begin{tabular}[c]{@{}c@{}} 89.58$\pm$1.33 \\ 68.48$\pm$0.88 \end{tabular} &
    \begin{tabular}[c]{@{}c@{}} 89.04$\pm$0.57 \\ 68.84$\pm$0.63 \end{tabular} &
    \begin{tabular}[c]{@{}c@{}} 99.55$\pm$0.70 \\ 65.41$\pm$9.87 \end{tabular} \\ [12pt]
    
    Ensemble & \textbf{\color{red}79.45} & \textbf{\color{red}731.71} & \textbf{\color{red}47.27} &
    \begin{tabular}[c]{@{}c@{}} 82.40 \\ \textbf{\color{red}75.54} \end{tabular} &
    \begin{tabular}[c]{@{}c@{}} 79.53 \\ \textbf{\color{blue}81.38} \end{tabular} &
    \multicolumn{1}{c}{\begin{tabular}[c]{@{}c@{}} 84.83 \\ \textbf{\color{blue}72.57} \end{tabular}} &
    \begin{tabular}[c]{@{}c@{}} 82.62 \\ 77.47 \end{tabular} &
    \begin{tabular}[c]{@{}c@{}} 34.73 \\ \textbf{\color{red}94.97} \end{tabular} \\ [12pt]
    
    AugMix & 75.77$\pm$0.22 & 752.18$\pm$3.32 & 65.94$\pm$0.89 & \begin{tabular}[c]{@{}c@{}} 86.81$\pm$0.93 \\ 73.12$\pm$0.09 \end{tabular} &
    \begin{tabular}[c]{@{}c@{}} 83.48$\pm$1.04 \\ 77.93$\pm$0.75 \end{tabular} &
    \begin{tabular}[c]{@{}c@{}} 87.12$\pm$0.87 \\ 71.05$\pm$0.68 \end{tabular} &
    \begin{tabular}[c]{@{}c@{}} 86.80$\pm$1.03 \\ 72.70$\pm$0.67 \end{tabular} &
    \begin{tabular}[c]{@{}c@{}} 57.52$\pm$39.02 \\ 88.93$\pm$9.08 \end{tabular} \\ [12pt]
    
    ODIN & 74.43$\pm$0.51 & 802.19$\pm$10.89 & 89.14$\pm$4.02 &
    \begin{tabular}[c]{@{}c@{}} 85.04$\pm$1.24 \\ 72.78$\pm$0.18 \end{tabular} &
    \begin{tabular}[c]{@{}c@{}} 88.38$\pm$1.58 \\ 74.98$\pm$0.96 \end{tabular} &
    \begin{tabular}[c]{@{}c@{}} 86.00$\pm$0.91 \\ 72.38$\pm$1.53 \end{tabular} &
    \begin{tabular}[c]{@{}c@{}} 87.07$\pm$2.20 \\ 70.56$\pm$1.97 \end{tabular} &
    \begin{tabular}[c]{@{}c@{}} 67.28$\pm$40.42 \\ 79.84$\pm$18.36 \end{tabular} \\ [12pt]
    
    OE & 72.85$\pm$0.61 & 739.73$\pm$12.99 & 93.63$\pm$6.05 &
    \begin{tabular}[c]{@{}c@{}} 88.33$\pm$0.61 \\ 71.07$\pm$0.70 \end{tabular} &
    \begin{tabular}[c]{@{}c@{}} 21.27$\pm$7.19 \\ \textbf{\color{red}96.12$\pm$1.42} \end{tabular} &
    \begin{tabular}[c]{@{}c@{}} 55.59$\pm$2.40 \\ \textbf{\color{red}86.93$\pm$0.92} \end{tabular} &
    \begin{tabular}[c]{@{}c@{}} 8.62$\pm$4.07 \\ \textbf{\color{red}97.74$\pm$0.93} \end{tabular} &
    \begin{tabular}[c]{@{}c@{}} 86.16$\pm$17.43 \\ 80.93$\pm$12.53 \end{tabular} \\ [12pt]

    Energy & 68.92$\pm$0.76 & 762.05$\pm$7.45 & 161.88$\pm$3.07 &
    \begin{tabular}[c]{@{}c@{}} 89.93$\pm$0.40 \\ 65.56$\pm$0.16 \end{tabular} &
    \begin{tabular}[c]{@{}c@{}} 33.89$\pm$4.93 \\ 93.86$\pm$1.29 \end{tabular} &
    \begin{tabular}[c]{@{}c@{}} 60.36$\pm$3.62 \\ 86.17$\pm$0.73 \end{tabular} &
    \begin{tabular}[c]{@{}c@{}} 30.72$\pm$3.37 \\ 93.05$\pm$0.89 \end{tabular} &
    \begin{tabular}[c]{@{}c@{}} 63.68$\pm$36.49 \\ 89.94$\pm$6.16 \end{tabular} \\ [12pt]

    ReAct & 74.44$\pm$0.46 & 801.93$\pm$33.23 & 90.08$\pm$3.58 &
    \begin{tabular}[c]{@{}c@{}} 86.14$\pm$1.06 \\ 72.37$\pm$0.18 \end{tabular} &
    \begin{tabular}[c]{@{}c@{}} 88.48$\pm$2.36 \\ 76.15$\pm$1.11 \end{tabular} &
    \begin{tabular}[c]{@{}c@{}} 87.28$\pm$0.79 \\ 71.67$\pm$1.25 \end{tabular} &
    \begin{tabular}[c]{@{}c@{}} 86.80$\pm$1.55 \\ 70.69$\pm$1.90 \end{tabular} &
    \begin{tabular}[c]{@{}c@{}} 99.11$\pm$1.57 \\ 55.84$\pm$23.24 \end{tabular} \\ [12pt]
    
    OpenMax & 74.01$\pm$0.34 & 768.89$\pm$9.64 & 162.61$\pm$12.19 &
    \begin{tabular}[c]{@{}c@{}} 91.97$\pm$0.71 \\ 64.15$\pm$0.40 \end{tabular} &
    \begin{tabular}[c]{@{}c@{}} 78.70$\pm$4.92 \\ 79.93$\pm$1.37 \end{tabular} &
    \begin{tabular}[c]{@{}c@{}} 91.17$\pm$0.69 \\ 71.18$\pm$1.52 \end{tabular} &
    \begin{tabular}[c]{@{}c@{}} 75.09$\pm$4.25 \\ \textbf{\color{blue}79.27$\pm$1.71} \end{tabular} &
    \begin{tabular}[c]{@{}c@{}} 54.15$\pm$47.77 \\ 82.99$\pm$16.49 \end{tabular} \\
    
    \bottomrule
    \end{tabular}%
    }
    \vskip -0.1in
\end{table*}

\subsection{CIFAR-based Benchmark} \label{sec:section4.1}
The performance comparison results on the CIFAR-based benchmark are summarized in Table~\ref{table:cifar-perf}. 
Overall, Deep Ensemble performs competitively with the best unknown detection performance. It shows better unknown detection performance than OE and Energy even though they exploit outliers during the training phase. This stems from the fact that Deep Ensemble achieves the highest classification accuracy on the in-distribution dataset, as expected, while OE and Energy lose $1.58\%$ and $5.51\%$ of their classification accuracy compared to the MCP, respectively.

The other methods for detecting OoD inputs, such as ODIN, ReAct, and OpenMax, show similar or even worse unknown detection performance relative to that of the MCP. As the MD performance of ODIN and OpenMax shows, we observe that this occurred because their approaches distort the softmax outputs of the in-distribution inputs to create a separation between the confidence scores of the in-distribution inputs and those of OoD inputs. Similarly, we infer that the reason for the accuracy loss and poor MD performance of OE and Energy is the side effect of learning the uniformly distributed outputs for auxiliary inputs. 

On the other hand, CRL and EDL, methods for MD, show better unknown detection performance than the MCP. Additionally, they generally outperform MCP in the in-distribution vs. near-/far-OoD setting. Although the improvement is not significant, this observation demonstrates the possibility that improving confidence estimates of in-distribution inputs positively affects the confidence estimates of the OoD inputs. AugMix shows competitive performance overall, which implies that training with appropriate regularization methods can improve the confidence estimates of both in-distribution and OoD inputs. MCdropout improves the MD performance of the MCP, but its performance on other tasks is generally worse than the MCP.

\begin{figure*}[!t]
    \subfigure[CIFAR-based benchmark]{
        \includegraphics[width=0.48\textwidth]{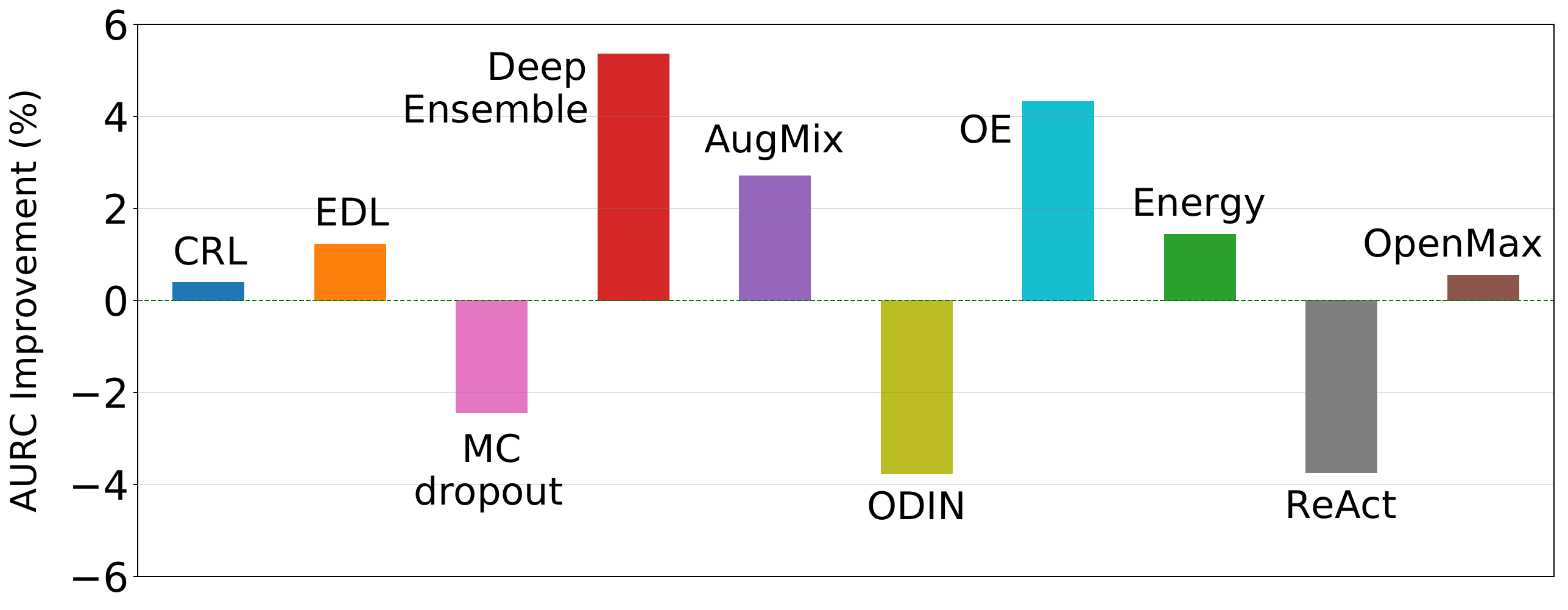}
        \label{fig:aurc-barh-a}
    }
    \subfigure[ImageNet-based benchmark]{
        \includegraphics[width=0.48\textwidth]{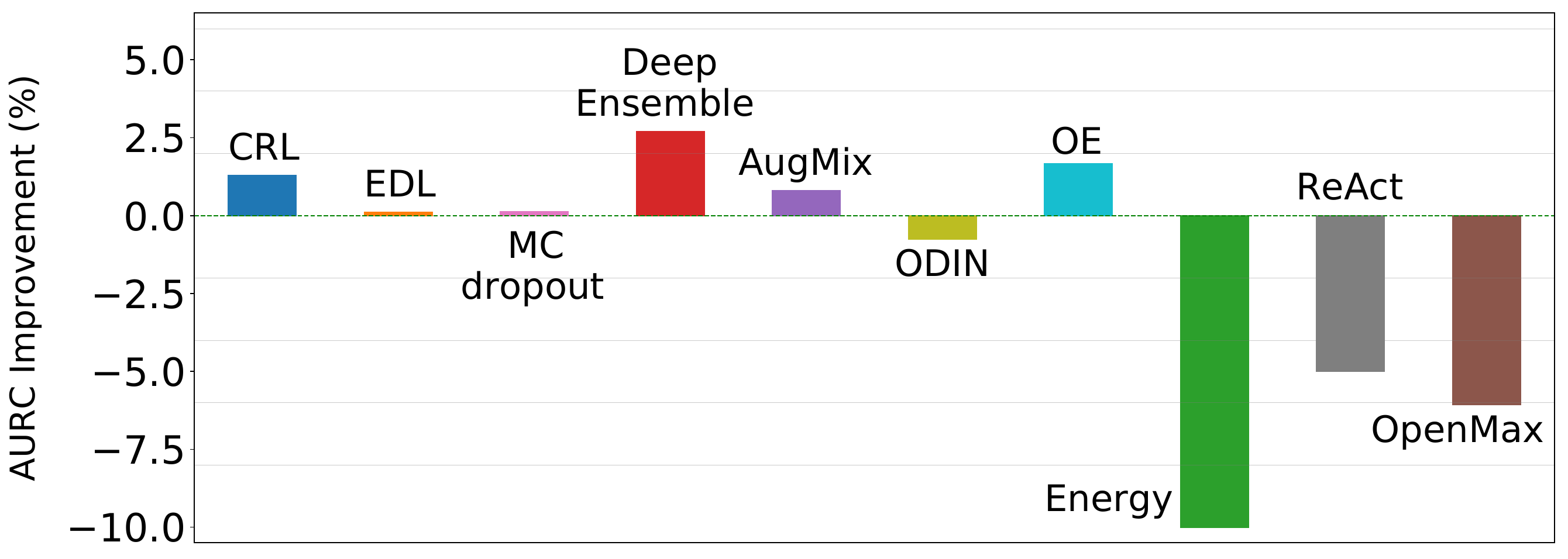}
        \label{fig:aurc-barh-b}
    }
    \caption{Performance improvement ratio of each comparison method in unknown detection. The center line ($0\%$, green dashed) indicates the MCP performance.}
    \label{fig:aurc-barh}
\vskip -0.1in
\end{figure*}

Figure~\ref{fig:aurc-barh-a} summarizes the improvement ratio of the unknown detection performance of the comparison methods over the MCP on the CIFAR-based benchmark. The methods of Deep Ensemble, AugMix, or OE show improved unknown detection performance compared to the MCP, but the other methods such as ODIN and ReAct have lower performance than even the MCP. These results clearly support the aim of our study: focusing only on a specific problem setting, MD or OoDD, hinders any effective and valid comparison of the unknown detection performance of deep neural networks.

It is worth noting that none of the comparison methods achieve a significant improvement on near-OoD detection, indicating that a close semantic distance between in-distribution and near-OoD causes the misalignment of the ordinal ranking between the confidence scores of the in-distribution inputs and near-OoD inputs, as discussed later in Section~\ref{sec:section4.3}.

\subsection{ImageNet-based Benchmark} \label{sec:section4.2}
We also examine the unknown detection performance of the comparison methods on the ImageNet-based benchmark, as summarized in Table~\ref{table:imagenet-perf}. In this benchmark, Deep Ensemble still outperforms the other comparison methods with the best classification accuracy and the best unknown detection performance. Unlike the results with the CIFAR-based benchmark, OE does not degrade the classification accuracy on the in-distribution dataset, and it increases the MD performance. This observation implies that the performance of OE is fairly sensitive to the dataset used as outliers during training, which limits the practical applicability of OE given that manual refinement of large-scale outliers is infeasible. On the other hand, Energy still suffers from a loss of accuracy and a decrease in unknown detection performance, and it also does not perform well in detecting OoD samples compared to MCP.

\begin{table*}[!t]
    \centering
    \caption{Comparison of unknown detection performance over the comparison methods on the ImageNet-based benchmark. The means and standard deviations over five runs are reported. $\downarrow$ and $\uparrow$ indicate that lower and higher values are better, respectively. \textcolor{red}{\textbf{Red}} represents the best performance among all methods, whereas \textcolor{blue}{\textbf{blue}} represents the best performance of the other methods except those trained on the auxiliary dataset (i.e., Outlier Exposure and Energy). The AURC values are multiplied by $10^3$, and the other values are percentages.}
    \label{table:imagenet-perf}
    \vskip 0.1in
    \resizebox{1.0\textwidth}{!}{%
    \begin{tabular}{l|c|c|c|c|ccc}
    \toprule
    
    \multicolumn{1}{c}{\textbf{\begin{tabular}[c]{@{}c@{}} \\ \end{tabular}}} &
    \multicolumn{1}{c|}{\textbf{\begin{tabular}[c]{@{}c@{}} \\ \end{tabular}}} &
    \textbf{\begin{tabular}[c]{@{}c@{}} Unknown \end{tabular}} &
    \textbf{\begin{tabular}[c]{@{}c@{}} Miscls. \end{tabular}} &
    \textbf{\begin{tabular}[c]{@{}c@{}} Near-OoD \end{tabular}} &
    \multicolumn{3}{c}{\textbf{\begin{tabular}[c]{@{}c@{}} Far-OoD Detection \end{tabular}}} \\ [3pt]
    
    \multicolumn{1}{c}{\textbf{\begin{tabular}[c]{@{}c@{}}  \end{tabular}}} &
    \multicolumn{1}{c|}{\textbf{\begin{tabular}[c]{@{}c@{}} \end{tabular}}} &
    \textbf{Detection} &
    \textbf{\begin{tabular}[c]{@{}c@{}} Detection \end{tabular}} &
    \textbf{\begin{tabular}[c]{@{}c@{}} Detection \end{tabular}} &
    \textbf{\begin{tabular}[c]{@{}c@{}} Food-32 \end{tabular}} &
    \textbf{\begin{tabular}[c]{@{}c@{}} Caltech-45 \end{tabular}} &
    \textbf{\begin{tabular}[c]{@{}c@{}} Places-82 \end{tabular}} \\ \hline
    
    \textbf{Method} & \textbf{ACC($\uparrow$)} & \multicolumn{2}{c|}{\textbf{AURC($\downarrow$)}} &
    \multicolumn{4}{c}{\textbf{\begin{tabular}[c]{@{}c@{}} FPR-95$\%$-TPR($\downarrow$) \\ AUROC($\uparrow$) \end{tabular}}} \\ \hline
    
    MCP & 81.09$\pm$0.65 & 585.74$\pm$7.13 & 42.12$\pm$4.79 &
    \begin{tabular}[c]{@{}c@{}} 82.28$\pm$1.28 \\ 77.51$\pm$0.44 \end{tabular} &
    \begin{tabular}[c]{@{}c@{}} 87.51$\pm$1.43 \\ 80.67$\pm$0.70 \end{tabular} &
    \begin{tabular}[c]{@{}c@{}} 69.90$\pm$1.50 \\ 83.10$\pm$2.14 \end{tabular} &
    \begin{tabular}[c]{@{}c@{}} 72.01$\pm$1.51 \\ 83.27$\pm$0.45 \end{tabular} \\ [12pt]

    CRL & 81.89$\pm$0.36 & 578.20$\pm$1.36 & 36.54$\pm$1.10 &
    \begin{tabular}[c]{@{}c@{}} 81.28$\pm$0.71 \\ 78.43$\pm$0.21 \end{tabular} &
    \begin{tabular}[c]{@{}c@{}} 88.12$\pm$0.67 \\ \textbf{\color{blue}80.86$\pm$0.34} \end{tabular} &
    \begin{tabular}[c]{@{}c@{}} 69.20$\pm$0.72 \\ 84.13$\pm$0.40 \end{tabular} &
    \begin{tabular}[c]{@{}c@{}} 72.90$\pm$0.99 \\ 83.34$\pm$0.35 \end{tabular} \\ [12pt]
    
    EDL & 77.97$\pm$1.48 & 585.15$\pm$5.15 & 46.32$\pm$4.95 &
    \begin{tabular}[c]{@{}c@{}} 86.77$\pm$1.04 \\ 75.92$\pm$0.65 \end{tabular} &
    \begin{tabular}[c]{@{}c@{}} 91.85$\pm$2.30 \\ 78.80$\pm$1.50 \end{tabular} &
    \begin{tabular}[c]{@{}c@{}} 75.75$\pm$1.53 \\ 83.71$\pm$0.85 \end{tabular} &
    \begin{tabular}[c]{@{}c@{}} 78.43$\pm$1.42 \\ 82.23$\pm$0.58 \end{tabular} \\ [12pt]
    
    MCdropout & 81.11$\pm$0.50 & 585.00$\pm$6.43 & 39.10$\pm$1.80 &
    \begin{tabular}[c]{@{}c@{}} 83.32$\pm$0.83 \\ 77.65$\pm$0.33 \end{tabular} &
    \begin{tabular}[c]{@{}c@{}} 89.79$\pm$0.87 \\ 79.99$\pm$0.49 \end{tabular} &
    \begin{tabular}[c]{@{}c@{}} 71.20$\pm$1.13 \\ 83.87$\pm$0.49 \end{tabular} &
    \begin{tabular}[c]{@{}c@{}} 73.78$\pm$1.23 \\ 83.18$\pm$0.32 \end{tabular} \\ [12pt]
    
    Ensemble & \textbf{\color{red}84.30} & \textbf{\color{red}569.98} & \textbf{\color{red}29.63} &
    \begin{tabular}[c]{@{}c@{}} 80.00 \\ \textbf{\color{blue}79.14} \end{tabular} &
    \begin{tabular}[c]{@{}c@{}} 88.03 \\ 80.70 \end{tabular} &
    \begin{tabular}[c]{@{}c@{}} 62.90 \\ 87.02 \end{tabular} &
    \begin{tabular}[c]{@{}c@{}} 69.54 \\ 84.85 \end{tabular} \\ [12pt]
    
    AugMix & 80.89$\pm$0.92 & 581.04$\pm$3.70 & 39.72$\pm$3.06 & 
    \begin{tabular}[c]{@{}c@{}} 83.90$\pm$0.61 \\ 77.23$\pm$0.39 \end{tabular} &
    \begin{tabular}[c]{@{}c@{}} 91.95$\pm$0.52 \\ 80.24$\pm$0.73 \end{tabular} &
    \begin{tabular}[c]{@{}c@{}} 70.25$\pm$1.15 \\ 84.69$\pm$0.49 \end{tabular} &
    \begin{tabular}[c]{@{}c@{}} 74.81$\pm$1.18 \\ 83.18$\pm$0.32 \end{tabular} \\ [12pt]
    
    ODIN & 81.09$\pm$0.65 & 590.13$\pm$6.36 & 45.33$\pm$4.62 &
    \begin{tabular}[c]{@{}c@{}} 82.53$\pm$1.16 \\ 77.39$\pm$0.27 \end{tabular} &
    \begin{tabular}[c]{@{}c@{}} 89.33$\pm$3.66 \\ 78.46$\pm$4.35 \end{tabular} &
    \begin{tabular}[c]{@{}c@{}} 68.81$\pm$3.52 \\ 83.54$\pm$2.82 \end{tabular} &
    \begin{tabular}[c]{@{}c@{}} 71.40$\pm$2.44 \\ 83.77$\pm$1.57 \end{tabular} \\ [12pt]
    
    OE & 83.21$\pm$0.43 & 575.96$\pm$6.89 & 34.37$\pm$4.47 &
    \begin{tabular}[c]{@{}c@{}} 79.35$\pm$1.28 \\ \textbf{\color{red}79.24$\pm$0.40} \end{tabular} &
    \begin{tabular}[c]{@{}c@{}} 88.18$\pm$0.93 \\ \textbf{\color{red}81.27$\pm$0.22} \end{tabular} &
    \begin{tabular}[c]{@{}c@{}} 65.14$\pm$1.15 \\ 84.85$\pm$2.19 \end{tabular} &
    \begin{tabular}[c]{@{}c@{}} 69.85$\pm$0.94 \\ 84.84$\pm$0.28 \end{tabular} \\ [12pt]

    Energy & 80.81$\pm$0.37 & 625.02$\pm$4.44 & 74.06$\pm$1.17 &
    \begin{tabular}[c]{@{}c@{}} 77.17$\pm$0.78 \\ 77.90$\pm$0.11 \end{tabular} &
    \begin{tabular}[c]{@{}c@{}} 97.48$\pm$0.41 \\ 63.51$\pm$1.56 \end{tabular} &
    \begin{tabular}[c]{@{}c@{}} 72.18$\pm$2.77 \\ 84.46$\pm$0.75 \end{tabular} &
    \begin{tabular}[c]{@{}c@{}} 89.39$\pm$1.15 \\ 77.69$\pm$0.36 \end{tabular} \\ [12pt]

    ReAct & 81.13$\pm$0.54 & 618.16$\pm$8.80 & 67.29$\pm$7.33 &
    \begin{tabular}[c]{@{}c@{}} 84.43$\pm$0.33 \\ 75.26$\pm$0.49 \end{tabular} &
    \begin{tabular}[c]{@{}c@{}} 95.72$\pm$0.37 \\ 68.19$\pm$1.45 \end{tabular} &
    \begin{tabular}[c]{@{}c@{}} 59.44$\pm$1.40 \\ 85.53$\pm$2.26 \end{tabular} &
    \begin{tabular}[c]{@{}c@{}} 66.45$\pm$0.80 \\ \textbf{\color{red}86.25$\pm$0.35} \end{tabular} \\ [12pt]
    
    OpenMax & 81.32$\pm$0.77 & 621.28$\pm$11.29 & 59.05$\pm$2.85 &
    \begin{tabular}[c]{@{}c@{}} 84.34$\pm$0.65 \\ 75.57$\pm$0.65 \end{tabular} &
    \begin{tabular}[c]{@{}c@{}} 95.21$\pm$0.57 \\ 67.52$\pm$1.15 \end{tabular} &
    \begin{tabular}[c]{@{}c@{}} 57.70$\pm$2.54 \\ \textbf{\color{red}87.96$\pm$0.55} \end{tabular} &
    \begin{tabular}[c]{@{}c@{}} 68.65$\pm$1.85 \\ 85.99$\pm$0.38 \end{tabular} \\
    
    \bottomrule
    \end{tabular}%
    }
    \vskip -0.1in
\end{table*}

ODIN and OpenMax perform well in near-/far-OoD detection as they target the detection of OoD inputs. However, their MD and unknown detection performance are worse than the MCP, as similarly observed in CIFAR-based benchmark results. This observation confirms again that increasing the gap between the confidence values of the in-distribution inputs and those of the OoD inputs by directly manipulating the softmax distribution is not a desirable approach in terms of unknown detection. ReAct also performs well on the Caltech-45 and Places-82 far-OoD datasets, but it shows worse MD and unknown detection performance than MCP.

CRL shows competitive unknown detection performance. There is no noticeable performance improvement in near-/far-OoD detection, but CRL produces good confidence estimates for in-distribution inputs, which positively affects the unknown detection performance.
AugMix shows improved performance in both unknown detection and MD from the MCP, but this improvement was not found for the detection of OoD inputs. This likely stems from that it is because of interpolated inputs created using diverse classes. 
Similar to the CIFAR-based benchmark results, MCdropout makes confidence estimates of in-distribution inputs much better than the MCP, but its performance on the other tasks are not much different from the MCP. EDL cannot easily classify numerous classes compared to the other methods compared here.

As in the results on the CIFAR-based benchmark, near-OoD is the most difficult task when seeking an improvement from the MCP. Although the ImageNet-based benchmark has far more diverse classes and far more images than the CIFAR-based benchmark, the models still have difficulty distinguishing the near-OoD inputs from the in-distribution inputs according to their confidence values.

Figure~\ref{fig:aurc-barh-b} illustrates the unknown detection improvement ratios of each method compared here over the MCP on ImageNet-based benchmark. It shows a tendency similar to that on the CIFAR-based benchmark overall. The methods of Deep Ensemble, AugMix, and OE are improved from the MCP, and ODIN and ReAct show low performance, as in the results on the CIFAR-based benchmark. Notably, the performance of Energy and OpenMax is significantly degraded on this benchmark, although OpenMax shows competitive far-OoD detection performance on Caltech-45 and Places-82. ReAct also detects far-OoD samples well, but has a substantial degradation in unknown detection performance.

\begin{figure*}[!t]
    \subfigure[MCP]{
        \includegraphics[width=0.45\textwidth]{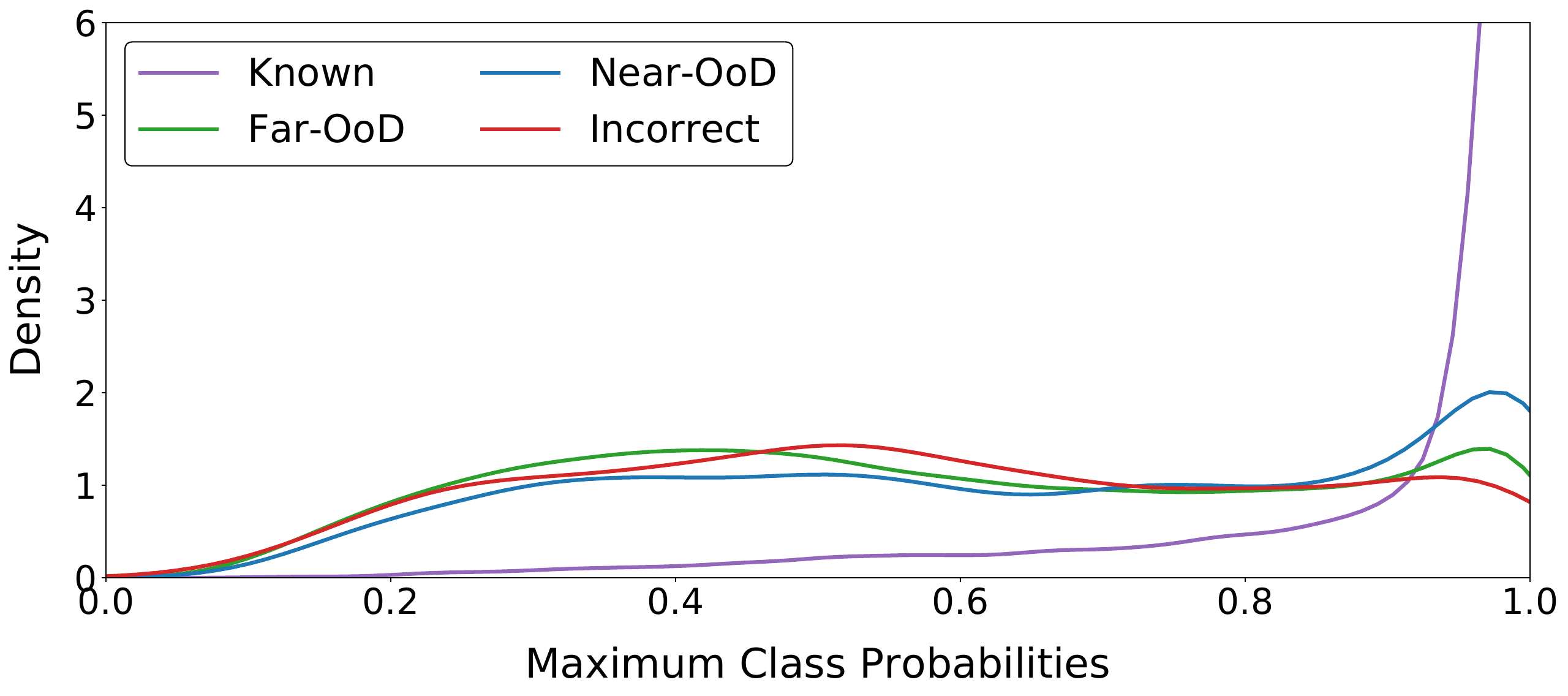}
        \label{fig:mcp-density-a}
    }
    \subfigure[Deep Ensemble]{
        \includegraphics[width=0.45\textwidth]{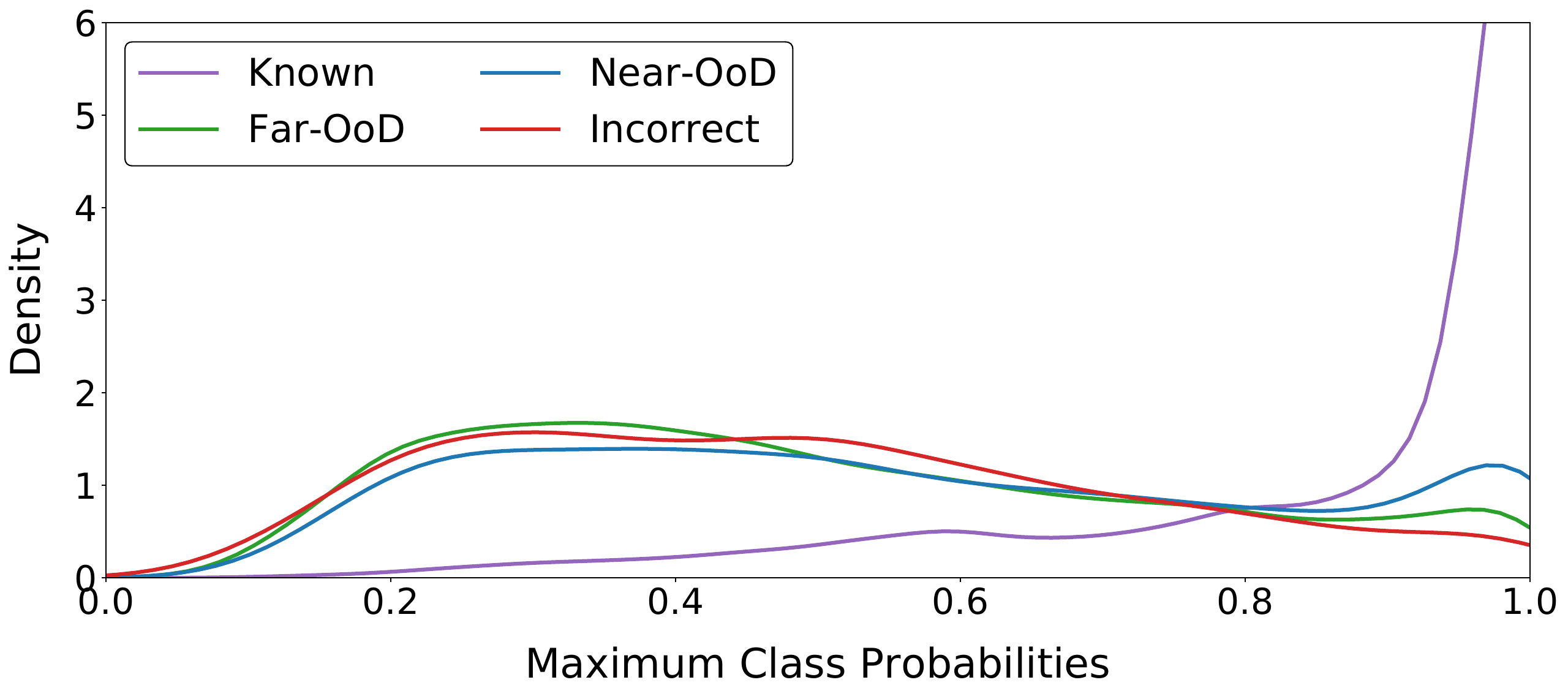}
        \label{fig:mcp-density-b}
    }
    \caption{Density plot of the maximum class probability for each category on the ImageNet-based benchmark.}
    \label{fig:mcp-density}
\vskip -0.1in
\end{figure*}

\begin{figure*}[!hp]
    \centering
    \subfigure[\scriptsize{CIFAR-60: MCP}]{
        \includegraphics[width=0.3\textwidth]{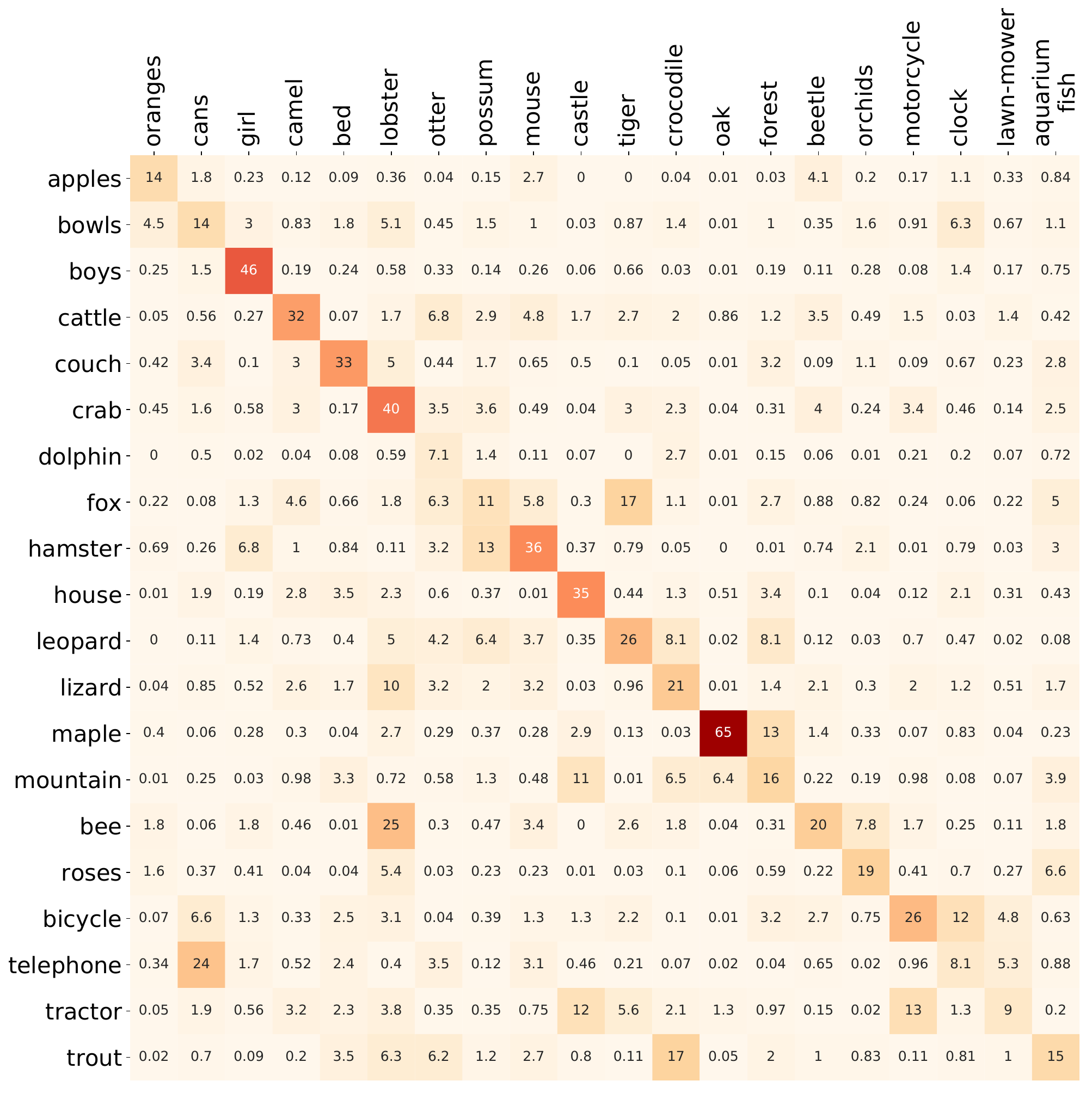}
        \label{fig:cifar-heatmap-baseline-near-a}
    }
    \subfigure[\scriptsize{CIFAR-60: Deep Ensemble}]{
        \includegraphics[width=0.3\textwidth]{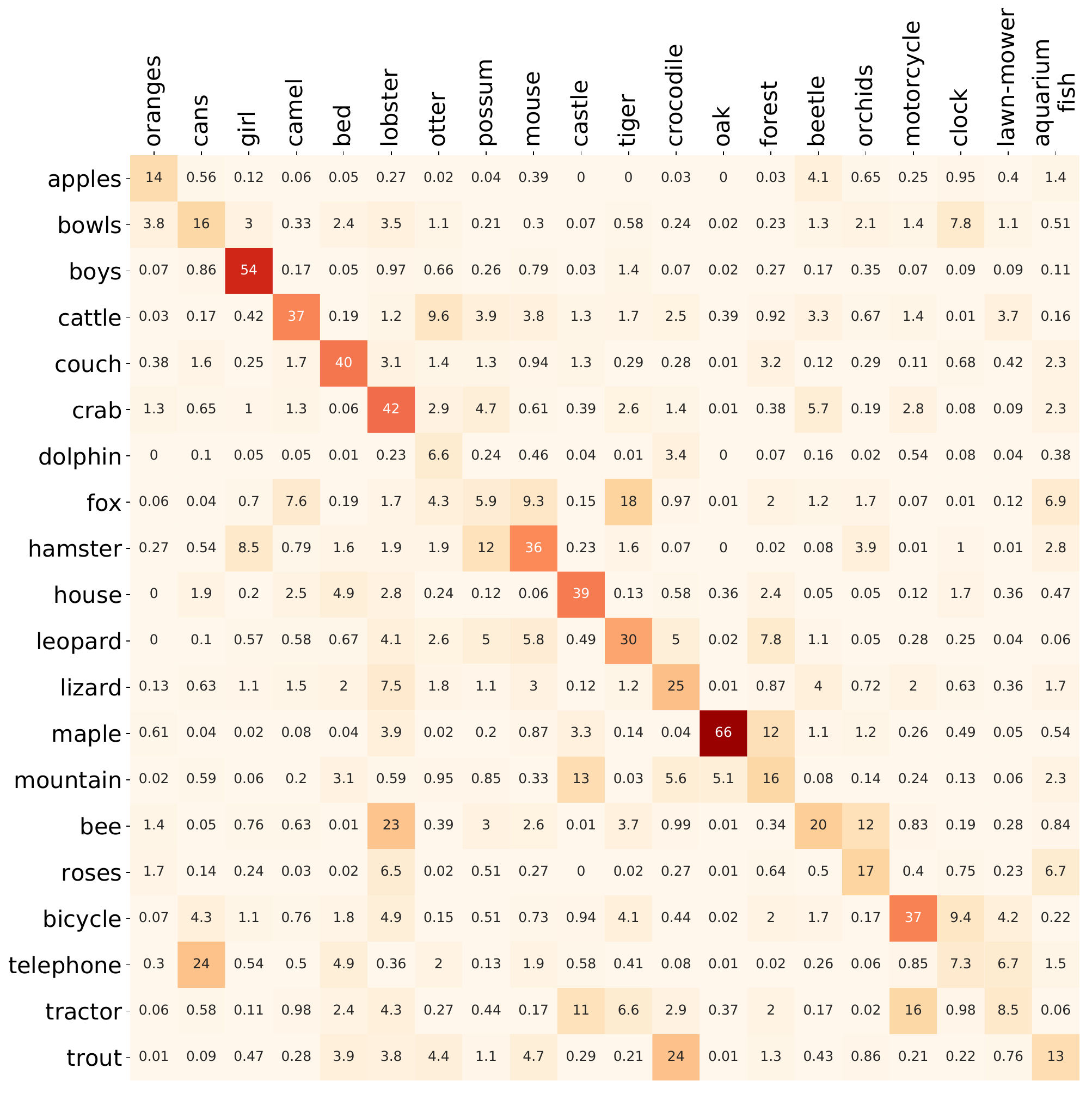}
        \label{fig:cifar-heatmap-ensemble-near-b}
    }
    \subfigure[\scriptsize{CIFAR-60: Outlier Exposure}]{
        \includegraphics[width=0.3\textwidth]{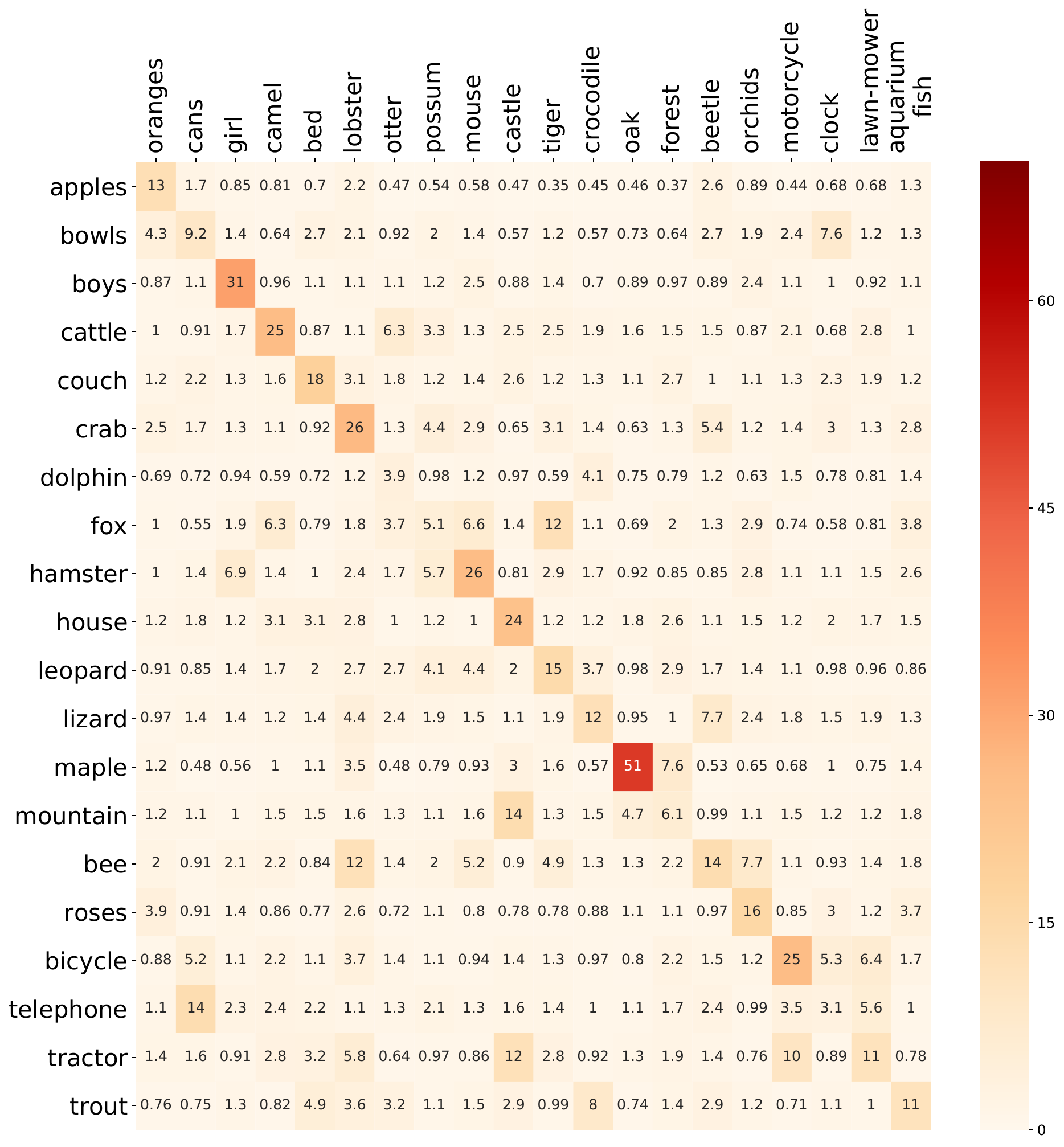}
        \label{fig:cifar-heatmap-oe-near-c}
    }
    \vskip 0 \baselineskip
    \subfigure[\scriptsize{SVHN: MCP}]{
        \includegraphics[width=0.3\textwidth]{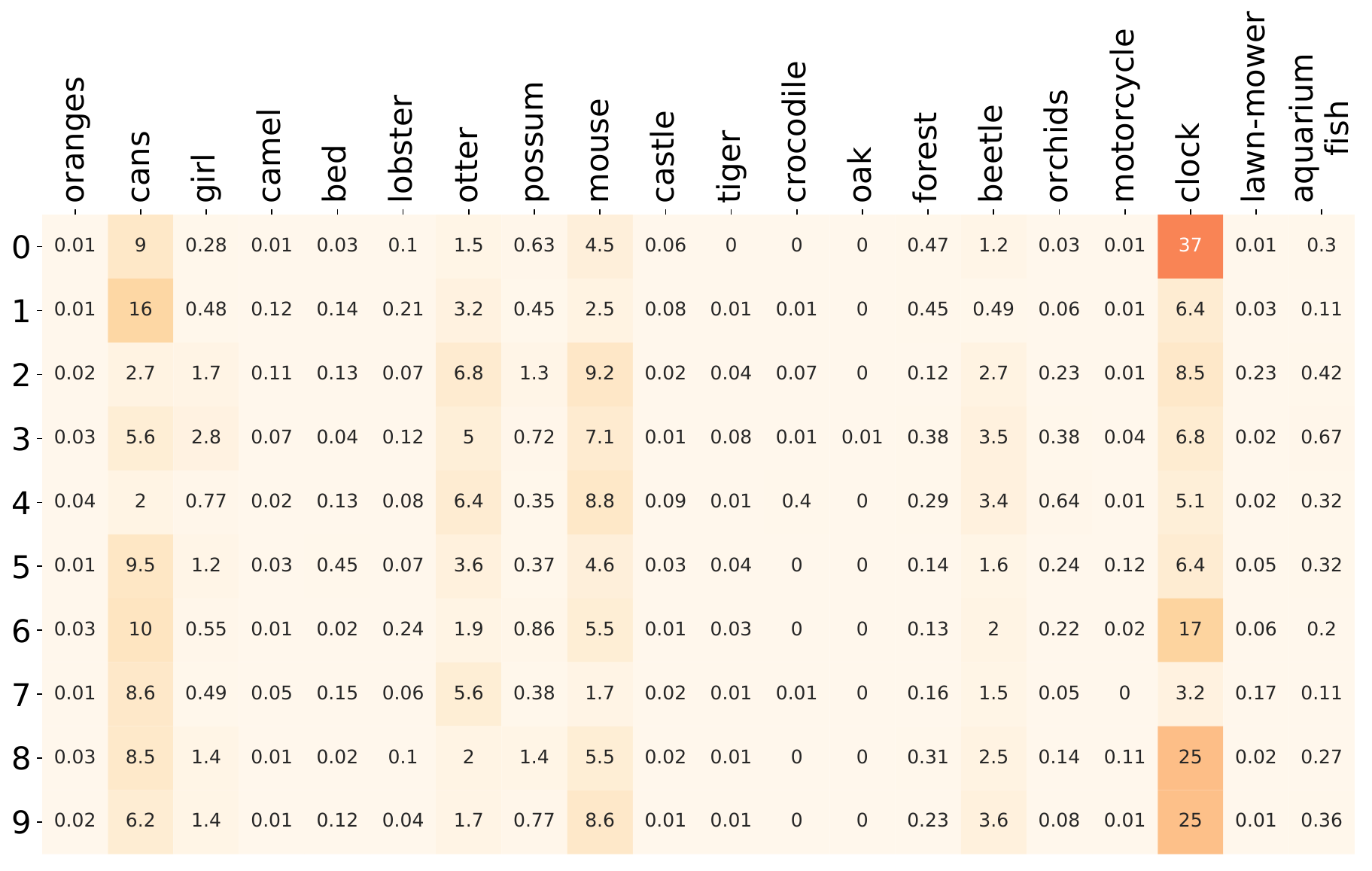}
        \label{fig:cifar-heatmap-baseline-svhn-d}
    }
    \subfigure[\scriptsize{SVHN: Deep Ensemble}]{
        \includegraphics[width=0.3\textwidth]{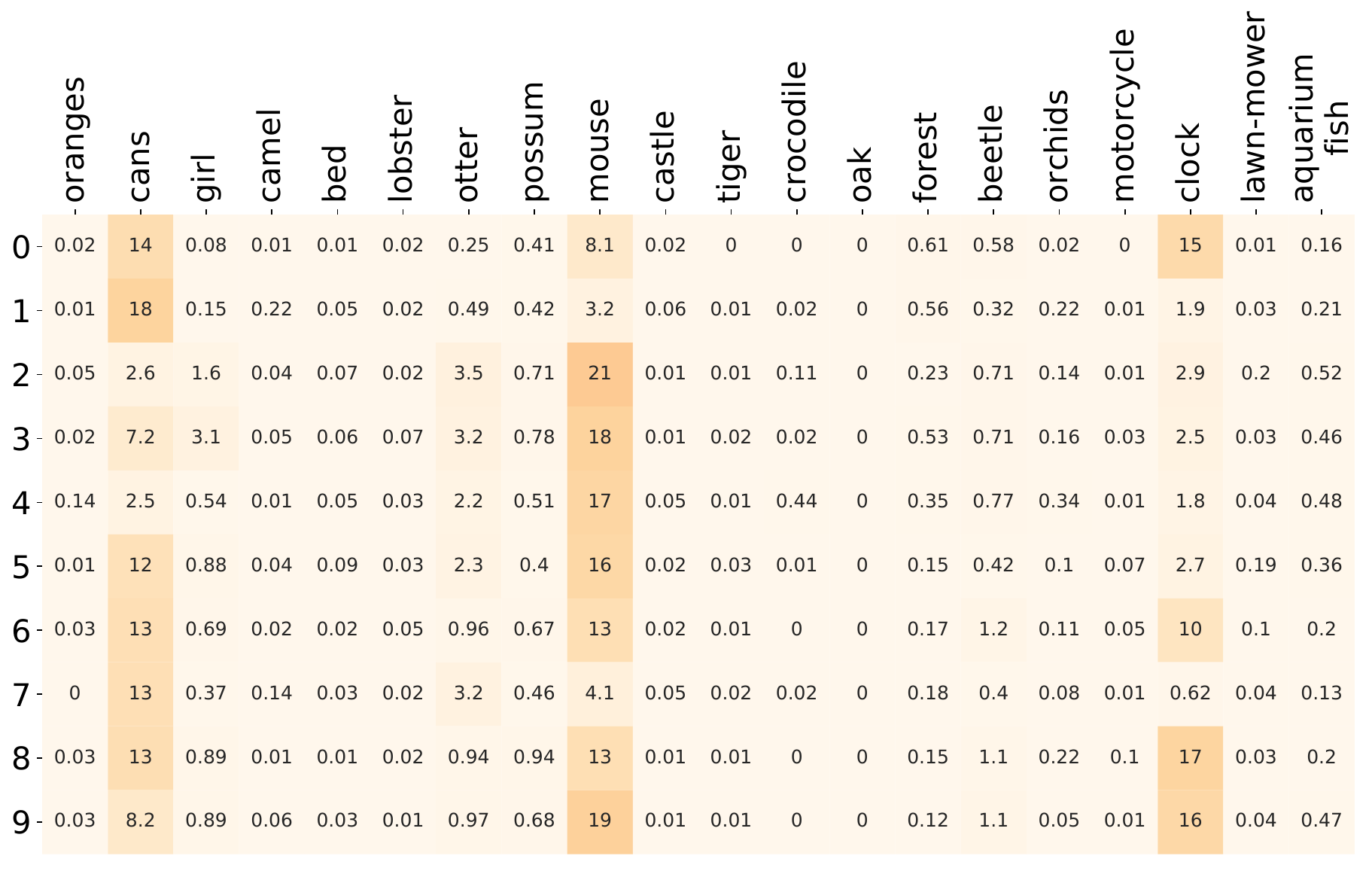}
        \label{fig:cifar-heatmap-ensemble-svhn-e}
    }
    \subfigure[\scriptsize{SVHN: Outlier Exposure}]{
        \includegraphics[width=0.3\textwidth]{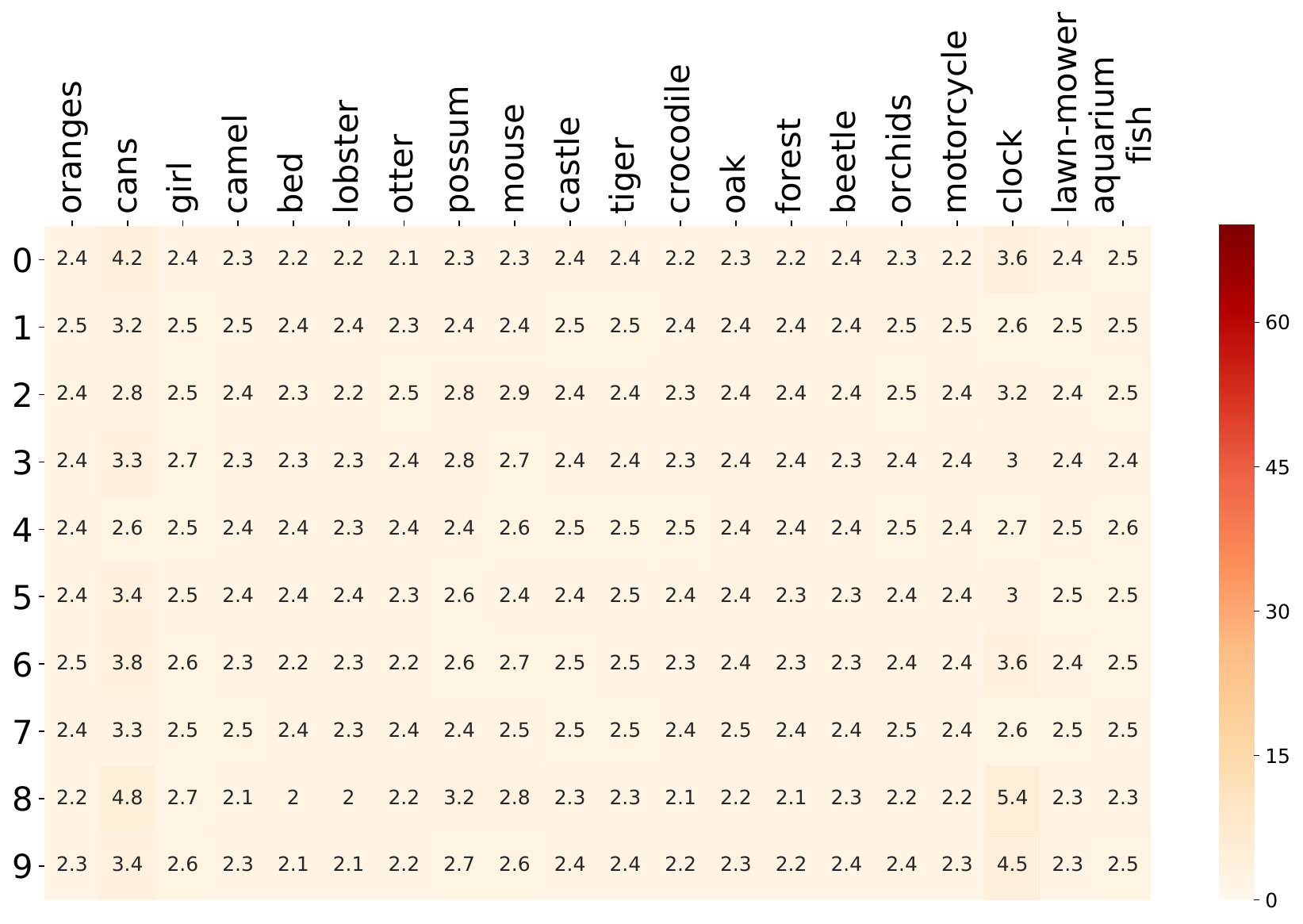}
        \label{fig:cifar-heatmap-oe-svhn-e}
    }
    \vskip 0 \baselineskip
    \subfigure[\scriptsize{DTD: MCP}]{
        \includegraphics[width=0.3\textwidth]{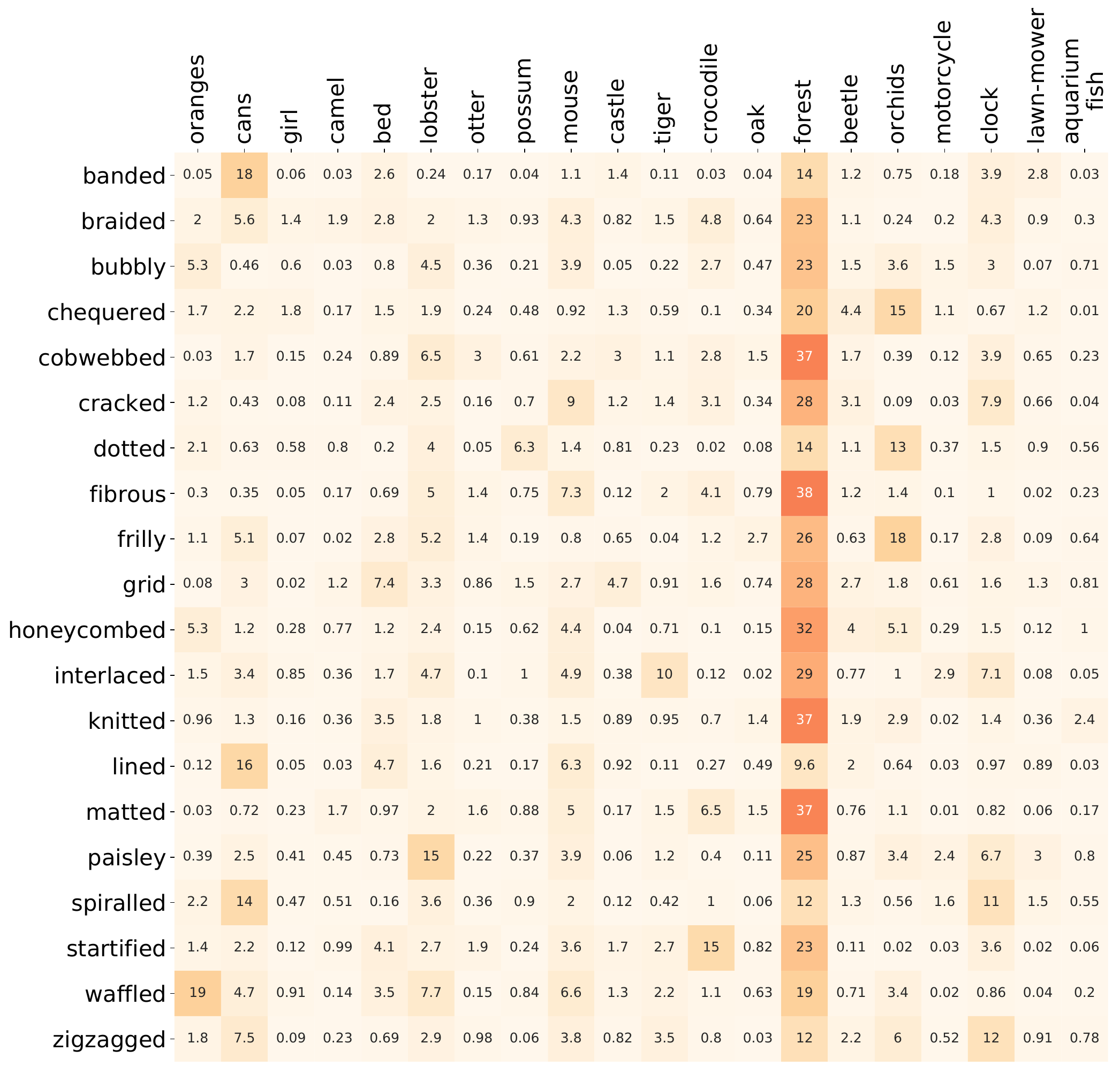}
        \label{fig:cifar-heatmap-baseline-dtd-f}
    }
    \subfigure[\scriptsize{DTD: Deep Ensemble}]{
        \includegraphics[width=0.3\textwidth]{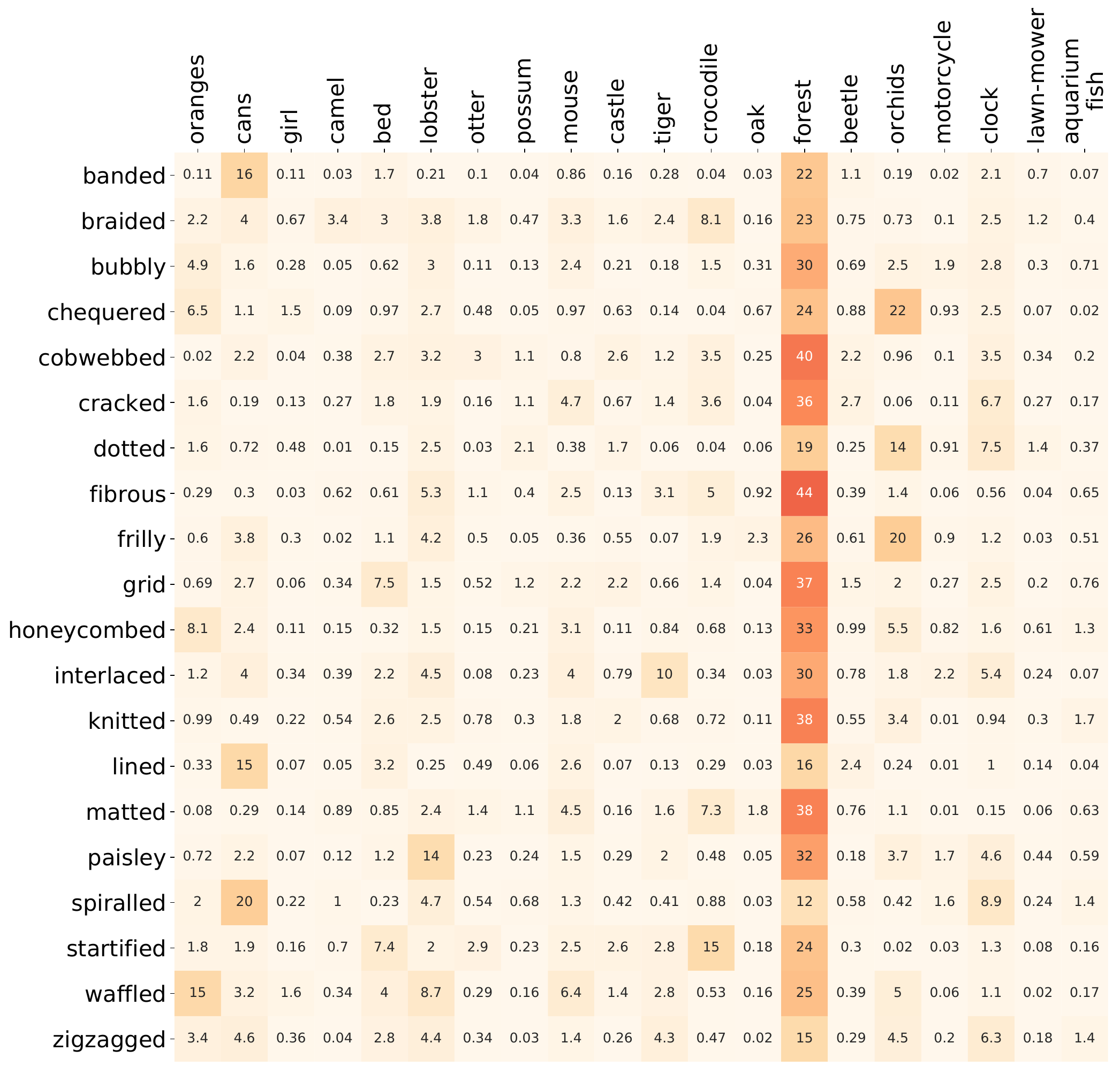}
        \label{fig:cifar-heatmap-ensemble-dtd-g}
    }
    \subfigure[\scriptsize{DTD: Outlier Exposure}]{
        \includegraphics[width=0.3\textwidth]{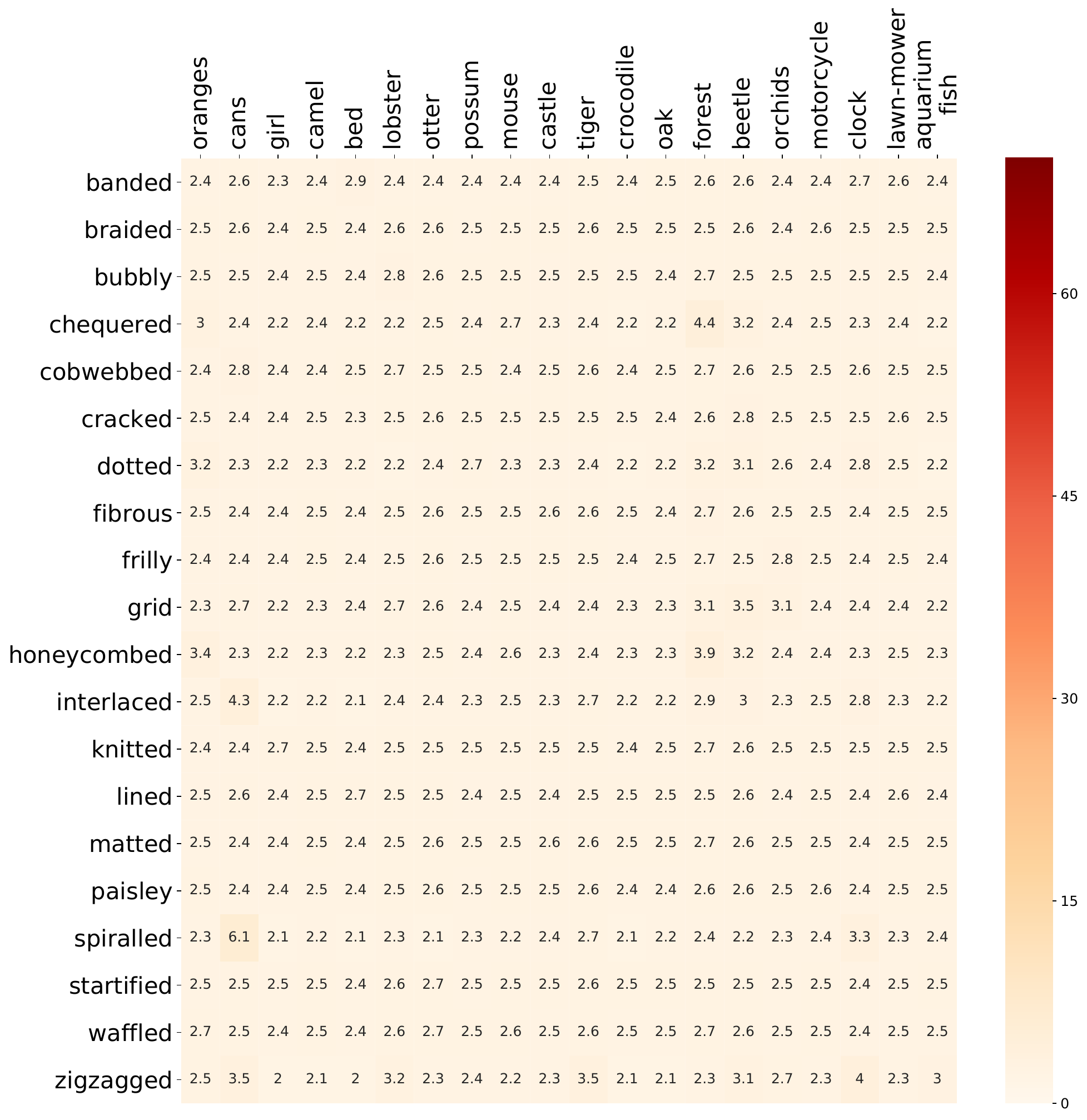}
        \label{fig:cifar-heatmap-oe-dtd-h}
    }
    \caption{Visualization of the softmax outputs of the MCP, Deep Ensemble, and Outlier Exposure on the CIFAR-based benchmark. The softmax values for a specific OoD class are averaged over all images in that class. The classes on each column in one heatmap consist of 20 classes from CIFAR-40 (in-distribution) sampled from each superclass. The classes on each row are sampled from the corresponding OoD dataset. For near-OoD cases, (a), (b), and (c), classes on the $i$-th column and row belong to the same superclass.}
    \label{fig:cifar-heatmap}
\vskip -0.1in
\end{figure*}

\subsection{Discussion} \label{sec:section4.3}
It is intuitively understandable that far-OoD inputs should have the lowest confidence values in their predictions among other input categories, meaning that known inputs and far-OoD inputs would be the most distinguishable. It is, however, not safe to assume which category should be more distinguishable from known inputs between incorrectly predicted inputs and near-OoD inputs, which belong to in-distribution and OoD, respectively.

Figure~\ref{fig:mcp-density} shows a density plot of the maximum class probabilities (MCP) produced by the MCP (Figure~\ref{fig:mcp-density-a}) and by Deep Ensemble (Figure~\ref{fig:mcp-density-b}) on the ImageNet-based benchmark. Near-OoD inputs generally have higher MCP than the far-OoD inputs. This observation can support the validity of the benchmark datasets that our near-OoD dataset is semantically closer to the in-distribution dataset compared to the far-OoD datasets. Deep Ensemble is more capable of detecting far-OoD inputs with MCP than the MCP, but it still assigns a slightly higher MCP to near-OoD inputs than incorrectly predicted inputs similar to the MCP. Specifically, on the ImageNet-based benchmark, AURC of the MCP for the known inputs vs. the near-OoD detection setting is $282.59\pm3.62$ (averaged over five runs, multiplied by $10^3$), which is much worse than AURC of the known inputs vs. incorrectly predicted inputs setting, at $42.12\pm4.79$. This tendency is also observed in the other comparison methods. We infer that this occurs because near-OoD inputs in our benchmarks share the high-level semantics with the in-distribution classes. 

Figure~\ref{fig:cifar-heatmap} visualizes the distribution of the softmax values from the MCP, Deep Ensemble, and OE for the near-/far-OoD classes in the CIFAR-based benchmark. The first row in this figure shows the softmax distribution associated with near-OoD classes (i.e. classes in CIFAR-60). All three models produce relatively high softmax values to the semantically close classes. On the other hand, all models produce widespread softmax probabilities for far-OoD classes, as demonstrated in the second and third rows in Figure~\ref{fig:cifar-heatmap}. Interestingly, the softmax distributions for far-OoD classes from DTD (the third row in the figure) show similar patterns: the probability mass is concentrated on a specific in-distribution class, \texttt{forest}. This implies that conventional deep softmax classifiers cannot be guaranteed to produce low predictive confidence for inputs far from the in-distribution. The third column in Figure~\ref{fig:cifar-heatmap} shows that OE generally produces uniformly distributed outputs for OoD inputs, as intended.

One noticeable observation is that the softmax outputs of Deep Ensemble for near-OoD classes do not differ much from those of the MCP, whereas this method outperforms the MCP on the near-OoD detection task, as shown in Table~\ref{table:cifar-perf}. This occurs because Deep Ensemble widens the gap of the confidence values between the in-distribution and OoD inputs by assigning higher softmax values to these confident predictions rather than reducing the confidence values of the near-OoD inputs. 

As described in Figure~\ref{fig:cifar-heatmap} and as shown by our experiment results, it is challenging to separate in-distribution inputs and near-OoD inputs based on confidence estimates from deep neural networks trained with a limited dataset. This is a task-specific problem, which means that not every task requires the detection of near-OoD inputs. For example, it is not necessary to detect new dog breeds as unknown classes if the target task is to classify dogs vs. cats. However, a model should consider new dog breeds as unknown classes if the task is to classify dog breeds. The detection capability of near-OoD inputs should be a great concern, especially for safety-critical applications such as autonomous driving. An autonomous car should identify a new traffic sign as unknown to prevent the occurrence of fatalities. This discussion leads us to important research topics: how can we train deep classifiers that have powerful detectability on a wide range of unknowns including near-OoD inputs? and is it possible to build a universal classifier whose confidence outputs can be adjusted by conditioning on a target task?

In terms of benchmark datasets, there could be inherent biases in datasets due to their data acquisition process or other hidden factors. Resized LSUN is one example of such bias, which significantly impacts the performance of OoDD, as evidenced by the overestimated detection performance on resized LSUN. Nevertheless, it is impossible to eliminate all inherent biases, so our far-OoD benchmarks might have the same limitations. However, near-OoD datasets are not subject to this limitation because they are constructed from the same dataset as the in-distribution datasets. By constructing far-OoD datasets from multiple sources, we can expect the effects of inherent biases to be averaged out.

\section{Conclusion}
In this paper, we highlight that measuring only specific detection capabilities is not enough to evaluate the confidence estimate quality of a deep neural network; hence, we define the \emph{unknown detection} task by integrating MD and OoDD to measure the true detection capability. Unknown detection aims to distinguish the confidence scores of unknown inputs, including incorrectly predicted inputs and OoD inputs, from those of known inputs that are correctly predicted. To evaluate the unknown detection performance, we propose unified benchmarks consisting of three categories: in-distribution, near-OoD and far-OoD. With the proposed benchmark datasets, we examine the unknown detection performance of popular methods proposed for MD and OoD inputs. Our experimental results demonstrate that the existing methods proposed for a specific task show lower performance than even the MCP on other tasks. Although Deep Ensemble shows the most competitive performance for unknown detection, it remains areas requiring improvements on detecting OoDD, especially on near-OoD inputs. We believe that the proposed unknown detection task and reconfigured benchmark datasets are valuable as a starting point for investigating the trustworthiness of deep neural networks.

\section*{Acknowledgement}
\noindent This work was supported by the National Research Foundation of Korea (NRF) grant funded by the Korea government (MSIT and the Ministry of Education) (NRF-2021R1C1C1011907 and NRF-2019R1A6A1A03032119).


\bibliography{cas-refs}





\end{document}